# Tensor-SIFT based Earth Mover's Distance for Contour Tracking

**Peihua Li**

**Abstract** Contour tracking in adverse environments is a challenging problem due to cluttered background, illumination variation, occlusion, and noise, among others. This paper presents a robust contour tracking method by contributing to some of the key issues involved, including (a) a region functional formulation and its optimization; (b) design of a robust and effective feature; and (c) development of an integrated tracking algorithm. First, we formulate a region functional based on robust Earth Mover's distance (EMD) with kernel density for distribution modeling, and propose a two-phase method for its optimization. In the first phase, letting the candidate contour be fixed, we express EMD as the transportation problem and solve it by the simplex algorithm. Next, using the theory of shape derivative, we make a perturbation analysis of the contour around the best solution to the transportation problem. This leads to a partial differential equation (PDE) that governs the contour evolution. Second, we design a novel and effective feature for tracking applications.

We propose a dimensionality reduction method by tensor decomposition, achieving a low-dimensional description of SIFT features called Tensor-SIFT for characterizing local image region properties. Applicable to both color and gray-level images, Tensor-SIFT is very distinctive, insensitive to illumination changes, and noise. Finally, we develop an integrated algorithm that combines various techniques of the simplex algorithm, narrow-band level set and fast marching algorithms. Particularly, we introduce an inter-frame initialization method and a stopping criterion for the termination of PDE iteration. Experiments in challenging image sequences show that the proposed work has promising performance.

# 1 Introduction

Contour tracking [3] has been an active research topic thanks to its capability to follow the deformable boundaries. It has widespread applications in various fields such as medical image processing, intelligent surveillance and human-machine interaction. The pioneering work of contour tracking can be traced back to Kass et al. [22]. They defined an energy functional of the object boundary and solved the functional

The research was supported by the National Natural Science Foundation of China (No. 60973080) and Key Project of Chinese Ministry of Education (No. 210063).

P. Li
School of Computer Science and Technology, Heilongjiang University, Harbin, China
E-mail: peihualj@hotmail.com

optimization problem using the calculus of variation (Euler-Lagragian equation). The resulting partial differential equation (PDE) was solved by the finite difference method that was iterated until convergence.

Over the past two decades, a great deal of literature has been published and considerable advances have been made in the field of contour tracking. In terms of the type of information used, contour tracking methods can be roughly divided into three classes: boundary-based [5, 22] which depends on image features along the object contour, region-based [39, 47] which prefers the information within the region enclosed by the object boundary, and those based on combination of the both [59]. From the perspective of contour representation, two kinds of methods can be classified: parametric contour methods [44] where the contours are explicitly represented by the parametric curves, and geometric contour approaches [5], where the contours are implicitly represented by zero level sets of some high dimensional functions [37]. The former often has fast convergence speed but provides less flexibility in handling curve topology, whereas the latter tends to be computationally expensive but can naturally handle topological deformation.

Despite the varying categorizations of and the different techniques used in contour tracking methods, three of the key issues are required to be researched: (a) how to formulate functionals and how to optimize them; (b) how to seek effective features for object representation; and (c) how to develop a robust and accurate tracking algorithm.

As regards functional formulation, the classical methods [5, 22] define the energy functionals that involve terms of image features and the smoothness constraint on the boundary curves. Other typical approaches are based on the Bayesian inference of region segmentation [10, 39], or based on the Mumford-Shah functionals [6, 46]. In recent years, a class of functionals has been presented independently that measures the similarity or distance between the target and candidate models [16, 20, 57]. The common methods for functional optimization use the calculus of variations to derive the corresponding PDEs. For some complex functionals involving region terms, deriving the corresponding PDEs via the calculus of variation is nontrivial because Green's theorem has to be used to transform the region functionals into boundary functionals [1]. However, using the theory of shape derivatives [50], we can deal with such region functionals in a straightforward and principled manner [1].

The Earth Mover's Distance (EMD) [48], known as the Wasserstein distance in mathematics [45], has been proven to be a robust distance measure between two distributions, outperforming commonly used probability measures such as Jeffrey divergence, $\chi^2$ statistics, and $L_1$ distance. Hence, in this paper, we formulate a region functional based on EMD with kernel density to represent the distributions for contour tracking. Optimizing this functional is nontrivial, and thus we propose a two-phase method for its optimization. In the first phase, assuming the candidate shape is fixed, we express the EMD as the transportation problem and solve it using the simplex algorithm [30]. In the second phase, using the theory of shape derivative, we make a perturbation analysis of the candidate contour around the best solution to the transportation problem. Thus, we derive the PDE that governs the contour evolution.

The features often used in contour tracking are image gradient, color, texture, optical flows, or interframe density. Earlier works have focused on gradients [5, 22] of the boundary delineating the object and the background. However, in cluttered images or texture images, the object boundary may be undistinguishable from the background. In addition, the image gradients are susceptible to noises present in the image. Probabilistic modeling of color is very common in the contour tracking papers [16]. As color features are sensitive to illumination variations, an online model update is usually required for robust tracking [31]. Under the assumptions of brightness constancy and smoothness [15], optical flows describing the apparent motion of appearance can be effectively used [18]. For videos captured by a still camera, inter-frame differences are useful for moving object tracking [38], whereas the global motion of the camera has to be estimated to compensate for the object motion [19] for videos captured by mobile cameras. For enhancing the discriminating power of image features, Brox et al. [4] investigated combination of color, texture, and motion in the active contour framework.

This paper proposes a novel and effective feature called tensor-SIFT for contour tracking application. SIFT features [28, 29] have proven to be very effective in describing local image characteristics. However, the high dimensionality of SIFT features limits its application in tracking fields. Hence, we introduce a tensor decomposition method for its dimensionality reduction. The resulting low dimensional feature, which is applicable to both color and gray-level images, is very distinctive and insensitive to illumination change and noise. The idea is that we regard SIFT features as a tensor (SIFT-As-Tensor), and thus tensor decomposition [23] can be applied which captures multi-factor (i.e.,two-dimensional spatial layout and phase histogram) relationships inherently present in the image data [53, 54]. In contrast, the methods of vectorizing SIFT features (SIFT-As-Vector) can only capture one-factor of the histogram information while losing the two-dimensional spatial layout in the data.

Based on the functional formulation and Tensor-SIFT described previously, we finally develop an integrated contour tracking algorithm. This algorithm combines multiple techniques intended for effective contour evolution, including the simplex algorithm [30], and the level set algorithm, and the re-initialization algorithm of fast marching [36, 49]. Particularly, we introduce an inter-frame initialization scheme. This scheme exploits the mean shift iteration to initialize the current frame by the tracking result at the previous frame. It can provide a more accurate initial contour and is helpful in avoiding possible local minimum in the functional. We also introduce a stopping criterion for terminating the PDE iteration.

The criterion fits the most recent EMD values with straight lines, and terminates the iteration when the line slope does not decrease any more. The iteration is also halted if the area variation between two consecutive frames is large, considering the spatio-temporal continuity.

The remainder of the paper is organized as follows. Section 2 begins with an overview of the literature related to the paper. Section 3 explores the novel feature of Tensor-SIFT. Section 4 formulates the region functional and proposes a two-phase method for its optimization. Section 5 introduces an integrated contour tracking algorithm and analyzes its computational cost. Section 6 presents the experiments and the corresponding discussions. Finally, the concluding remarks are given in section 7.

## 2 Related Work

This section reviews the studies that are related to this paper, including sudies on the EMD, on SIFT features, and on tensor decomposition.

### 2.1 Transportation Problem and EMD

The transportation problem, also known as the Monge-Kantorivich problem, can be traced back to the 18th century [33]; it was reformulated and extended by Kantorovich [21] as a minimal distance problem (called Wasserstein distance) between two probability measures. Over the past years, it has been widely investigated and has found a variety of applications in many fields [45]. The discrete transportation problem is well studied in linear programming and can be solved by the simplex algorithm [30].

One of the first papers published is Peleg et al. [43], which introduced the transportation problem into image processing. The authors measured the distance between two gray-level images for image matching. Haker et al. [17] computed image registration and warping maps based on the Monge-Kantorovich theory of optimal mass transport and developed an efficient PDE for solving the Wasserstein distance. Rubner et al. [48] proposed a cross-bin probability measure, EMD, for distributions matching in content-based image retrieval. They showed the distinct advantages of the EMD over the other commonly used point-wise probability measures [48]. As EMD computation in high dimensional cases is computationally expensive, recent research efforts are devoted to presenting new EMD variants and/or to developing fast algorithms for EMD computation [26, 41, 42]. In the one-dimensional case the closed-form solution of EMD (Wasserstein distance) is used in a region-based contour model for image segmentation [35]. Application of EMD to object tracking was studied by Zhao et al. [58]. In this study, by describing the object shapes with ellipses, they adopted EMD to compare the reference and candidate models, and proposed an efficient gradient descent method called differential EMD (DEMD) to estimate the translation of elliptical objects.

Our work is similar to that of Ni et al. [35] in the sense that both nonparametric density and region-based Wasserstein distance are used in the active contour framework. The main difference is that in a multiple-dimensional case such as ours, the closed-form solution of the EMD does not exist, and therefore the method of Ni et al. [35] is not applicable to our problem. Our work is also motivated by [58]; the primary distinction is that our focus is on how to follow the complex contour shape and non-rigid deformation rather than the simple elliptical shapes with only the motion of translation. This naturally leads to a different paradigm–functional formulation and its optimization through PDE. Establishing region functionals by probability similarity or distance measures has also been investigated by previous studies [16, 20, 57]. However, our work differs from these in the probability measures adopted: we argue that for contour tracking problems EMD is more robust than other commonly used point-wise probability measures.

## 2.2 SIFT Features

The SIFT algorithm includes two main steps: keypoint detection and keypoint description [28, 29]. The first step consists of local extrema (called keypoints) detection, followed by keypoints localization and the dominant orientation determination. In this way, generally sparse keypoints are identified. In the second step, each keypoint is represented by phase histograms of pixel gradient magnitudes (hereafter called SIFT feature) in the image patch centered at the keypoint. The SIFT algorithm has demonstrated successful applications in object recognition [28, 34], image classification [11], image retrieval [56], and so on.

On the other hand, dense SIFT features computed at regular grid points, without the preceding step of keypoint detection, have also shown promising performance in scene category recognition [24, 25] or in aligning images of complex scenes [27]. Because SIFT features are of high dimensionality, it is desirable to get a low-dimensional description for both computational efficiency and feasible probabilistic modeling (e.g., histogram). In the bag-of-feature literature, the common practice is to reduce the dimensionality of SIFT feature by creating codewords through clustering algorithms followed by index histogram [24, 25]. Dimensionality reduction of SIFT features by PCA is also a natural choice [27]. Rather than using the SIFT descriptor proposed by Lowe [29], Yan and Sukthankar [56] described the characteristics of an image patch by concatenating into a vector the horizontal and vertical gradients of each image pixel in this patch and then used PCA for dimensionality reduction.

However, a SIFT feature is actually a $4{\times}4{\times}8$ histogram array of gradient magnitudes; its locally two-dimensional spatial information inherently present is lost when packed as a 128-element vector [11, 25, 27–29, 32, 34]. This analysis motivates us to exploit tensor decomposition for dimensionality reduction of SIFT features, which distinguishes our work from the previous ones. As tensor decompo-sition method can capture multi-factor relationships inherently present in data, the proposed low- dimensional feature, Tensor-SIFT, is more powerful in capturing both spatial layout and histogram information in describing the patch characteristics.

### 2.3 Tensor Decomposition

A tensor is a high-dimensional array, and can be considered a generalization of a vector (first-order tensor) and a matrix (second-order tensor). Tensor decomposition is one of the topics in multi-linear algebra, which reveals that a higher-order tensor is formed by a confluence of multiple factors, and its decomposition consists in exploring these multiple factors inherently present in data. Vasilescu and Terzopoulos [52] proposed a method of *TensorFaces* for face recognition involving multiple factors, such as facial geometry, expression, pose, and illumination. They used a technique of N-modes SVD, a multilinear extension to the classical matrix SVD, achieving significantly better recognition rates over the classical method of *eigenfaces* [51], which is dependent on linear PCA. Based on N-modes SVD, they also demonstrated effective dimensionality reduction in facial image ensembles [53]. Wang and Ahuja [54] presented a dimensionality reduction approach based on tensor-decomposition for effectively capturing the spatial and temporal redundancies. Their method achieved the most compact data representations among the state of the art [54].

## 3 Tensor-SIFT: Dimensionality Reduction of SIFT Features by Tensor Decomposition

In this section, we first interpret the SIFT feature from two different views: SIFT-As-Tensor and SIFT-As-Vector. We then describe the method of tensor-decomposition for dimensionality reduction of SIFT feature, achieving our novel feature of Tensor-SIFT.

### 3.1 SIFT Feature as a Multi-Dimensional Array

The computation of SIFT is described briefly as follows [refer to Lowe [29] for details]. First, for the interest point, we determine an $8\times8$ sampling image window centered at this point and then compute the gradient magnitude weighted by a Gaussian function and phase of every pixel. Next, the sampling window is regularly divided into $4\times4$ sub-windows; in each subwindow $[0, 2\pi]$ is uniformly partitioned into eight intervals, and the corresponding phase histogram is computed. To avoid boundary effects, trilinear interpolation is exploited to distribute the value of each gradient sample into the neighboring phase bins.

Fig. 1(a) shows an $8\times8$ image window. Fig. 1(b) presents the corresponding phase histograms computed in this window, where the arrow direction is the histogram bin indicator, and its length signifies the histogram bin value. Traditionally, this 4x4 histogram array is packed into a vector of 128 as shown in Fig. 1(c). As described previously, one SIFT feature describes the local region characteristics associated with the center pixel, including both the subwindows' spatial layout and their phase histograms. Vectorization of SIFT feature (SIFT-As-Vector) results in the loss of spatial layout inherently present in data. Clearly, the data of $4\times4\times8$ array are produced by the confluence of multiple factors, i.e., 2D spatial layout and histogram. Thus, the best way is to see the data "as is". This naturally leads to our view of "SIFT-As-Tensor": one SIFT feature is a third-order tensor, and a set of SIFT features is thus a fourth-order tensor.

### 3.2 Dimensionality Reduction by Tensor Decomposition

An $N$th-order or $N$-way tensor $\mathfrak{X}$ is an $N$-dimensional array. Mathematically, it is an element of the direct product of $N$ vector spaces, i.e., $\mathfrak{X} \in \mathbb{R}^{I_1 \times I_2 \times \cdots \times I_N}$, where $I_n$, $n = 1, \ldots, N$, denotes the size of dimension $n$. A tensor is rank-one if it can be written as the outer product of $N$ vectors. For easy mathematical manipulation a tensor is often flattened or unfolded into a matrix. The mode-$n$ matrix of a $N$-way tensor is a $I_n$ by $(I_1 \cdots I_{n-1} I_{n+1} \cdots I_N)$ matrix $\mathbf{X}^{(n)}$, where the tensor element indexed by $(i_1, i_2, \ldots, i_N)$ is mapped to the matrix element indexed by $(i_n, j)$ where

$$j = 1 + \sum_{\substack{k=1 \\ k \neq n}}^{N} (i_k - 1) \prod_{\substack{m=1 \\ m \neq n}}^{k-1} I_m$$

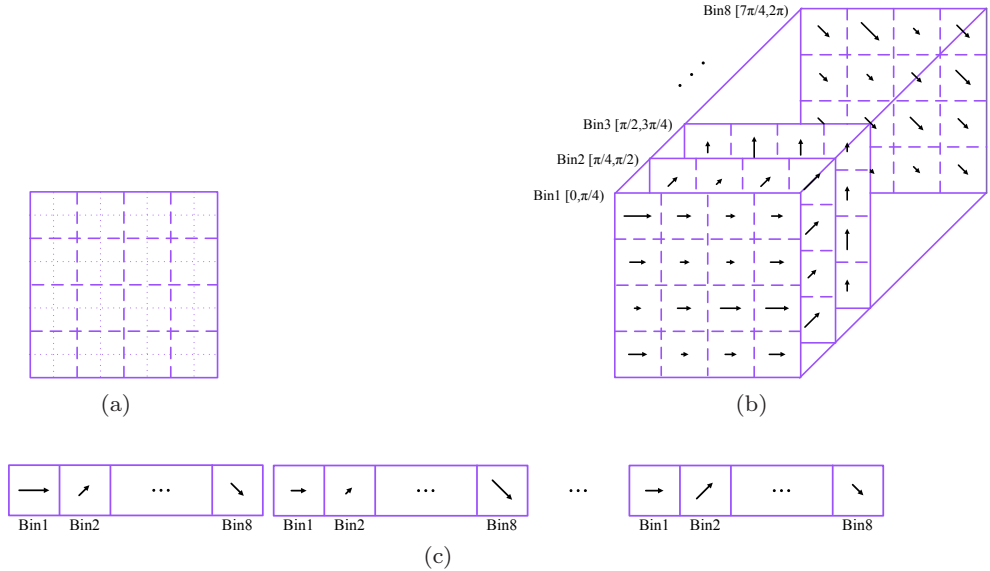

**Fig. 1** Illustration of one SIFT feature as a multi-dimensional array *vs* as a vector. An 8 by 8 sampling image window (a) is regularly divided into 4 by 4 sub-windows. For each sub-window an 8-bin phase histogram of gradient magnitudes is computed, producing a three-dimensional, 4 by 4 by 8, array (b), where the arrow direction is the histogram bin indicator, and its length signifies the histogram bin value. Traditionally, this three-dimensional array is vectorized into a one-dimensional vector of 128 (c), resulting in the loss of two-dimensional spatial layout inherently present in the data.

The idea of CANDECOMP/PARAFAC (CP) decomposition [13] is to approximate a tensor with a sum of component rank-one tensors. In our case, a set of SIFT features $\mathcal{F}$ is a fourth-order tensor, $\mathcal{F} \in \mathbb{R}^{I_1 \times I_2 \times I_3 \times I_4}$, where $I_1$ is the number of SIFT features, $I_2 = 4$, $I_3 = 4$ denote the numbers of horizontal and vertical sub-windows in the sampling image window, respectively, and $I_4 = 8$ denotes the histogram bin number. Its CP decomposition can be described by

$$\arg\min_{\hat{\mathcal{F}}} \| \mathcal{F} - \hat{\mathcal{F}} \|$$
$$\hat{\mathcal{F}} = \sum_{k=1}^{K} \mathbf{f}_k \circ \mathbf{r}_k \circ \mathbf{s}_k \circ \mathbf{t}_k \qquad (1)$$

where $\| \cdot \|$ denotes the tensor norm, i.e., the sum of the square of each element in the tensor, $\circ$ denotes outer product, $\mathbf{f}_k \in \mathbb{R}^{I_1}$, $\mathbf{r}_k \in \mathbb{R}^{I_2}$, $\mathbf{s}_k \in \mathbb{R}^{I_3}$, $\mathbf{t}_k \in \mathbb{R}^{I_4}$, and $K$ is the number of component rank-one tensors.

Denoting $\mathbf{F} = [\mathbf{f}_1 \ \ldots \ \mathbf{f}_K]$ and likewise for $\mathbf{R}$, $\mathbf{S}$ and $\mathbf{T}$, the minimization problem becomes how to seek the above four matrices that satisfy (1). We use alternating least square (ALS) algorithm [23] for its solution; i.e., we alternatively fix three of the four matrices, solving for the remaining one. We repeat this procedure until the error is less than a threshold or the maximum number of iterations is reached. The problem is reduced to a least square problem when fixing all but one matrix. For instance, let $\mathbf{R}$, $\mathbf{S}$, and $\mathbf{T}$ be fixed. The minimization problem is reduced to the least square problem in matrix form:

$$\arg\min_{\hat{\mathbf{F}}} \| \mathbf{F}^{(1)} - \hat{\mathbf{F}} (\mathbf{R} \odot \mathbf{S} \odot \mathbf{T})^T \|_F \qquad (2)$$

where $\mathbf{F}^{(1)}$ is the mode-$n$ matrix of the tensor $\mathcal{F}$, $\| \cdot \|_F$ stands for the matrix Frobenius norm, and $\odot$ denotes the Khatri-Rao product of two matrices. The Khatri-Rao product is defined as the column-wise Kronecker product of two matrices, e.g., $\mathbf{R} \odot \mathbf{S}$ is a matrix of $I_2 I_3$ by $K$:

$$\mathbf{R} \odot \mathbf{S} = [\mathbf{r}_1 \otimes \mathbf{s}_1 \ \ldots \ \mathbf{r}_K \otimes \mathbf{s}_K]$$

where $\otimes$ denotes the kronecker product. The solution to (2) can be given conveniently by the formula

$$\mathbf{F} = \mathbf{F}^{(1)} \big[ (\mathbf{R} \odot \mathbf{S} \odot \mathbf{T})^T \big]^\dagger \qquad (3)$$

where $\dagger$ denotes the Moore-Penrose pseudo-inverse.

Given a reference image, we compute the SIFT feature for every image pixel and arrange all SIFT features into a four-dimensional array, or a four-way tensor, $\mathcal{F} \in \mathbb{R}^{I_1 \times I_2 \times I_3 \times I_4}$. After performing its CP decomposition by the ALS algorithm described previously, the $I_1$ by $K$ matrix $\mathbf{F}$ can be considered the reference Tensor-SIFT of reduced dimensionality, and $I_2$ by $K$ matrix $\mathbf{R}$, $I_3$ by $K$ matrix $\mathbf{S}$ as well as $I_4$ by $K$ matrix $\mathbf{R}$ can be regarded as the tensor "basis matrices". When a new image is available, we compute the dense SIFT features for every pixel; each SIFT feature is a $1 \times 4 \times 4 \times 8$ tensor that can be approximated by projection (3). The resulting 1 by $K$ vector is the corresponding Tensor-SIFT of reduced dimensionality. Fig. 2 shows an example of the tensor decomposition ($K = 3$), where Fig. 2(a) shows a reference image (left), the scatter map of its tensor-SIFT (middle), and the tensor-SIFT image (right). Fig 2(b) shows four images (top row) and the corresponding tensor-SIFT images (bottom row). Note that the values of the Tensor-SIFT are scaled to $[0, 255]$ for each dimension.

## 4 Region Functional Formulation and Its Optimization

This section begins with the formulation of the region functional based on EMD (section 4.1) and then introduces a two-phase method for the functional optimization: the first-phase for solving EMD by the simplex algorithm (section 4.2) and the second for PDE derivation by shape derivative (section 4.3).

### 4.1 Formulation of the Region Functional

Following Rubner et al. [48], the reference and candidate models are denoted by $Signatatures$ $(\mathbf{h}_u^*, p_u)$ for $u = 1, \cdots, U$ and $(\mathbf{h}_v, q_v)$ for $v = 1, \cdots, V$, respectively, where $\mathbf{h}_u^*$ $(\mathbf{h}_v)$ and $p_u$ $(q_v)$ denote the $u$th subspace center and the corresponding distribution in the reference (candidate) object feature space. The distance between the reference and candidate models may be interpreted as the classical transportation problem that can be solved by the simplex algorithm.

There are many methods for distribution modeling, either parametric or non-parametric. In this paper we use the non-parametric kernel method [8] because it does not assume any prior distribution of the data and considers the (weak) spatial information. Suppose that the feature space of the reference object is divided into $U$ subspaces (or histogram bins), and the object is described by a complex image region $\Omega^*$ with a boundary $\Gamma^*$. Suppose that the object center is at the coordinate origin. Let $\mathbf{z}_i^* = [x_i^* \ y_i^*]^T$ be a point in the region $\Omega^*$ with feature vector $\mathbf{I}(\mathbf{z}_i^*)$, the reference distribution $p_u$ for $u = 1, \cdots, U$ of the object can be represented by

$$p_u = \frac{1}{\sum\limits_{\mathbf{z}_i^* \in \Omega^*} w(\mathbf{z}_i^*)} \sum\limits_{\mathbf{z}_i^* \in \Omega^*} w(\mathbf{z}_i^*) \delta_u(\mathbf{I}(\mathbf{z}_i^*)) \qquad (4)$$

where $w(\cdot)$ is a kernel function, and $\delta_u(\mathbf{I}(\mathbf{z}_i^*))$ denotes the Kronecker delta function that equals 1 if $\mathbf{I}(\mathbf{z}_i^*)$ belongs to the $u$th subspace, and otherwise equals 0. Note that the kernel function weights the point's contribution: one point that is closer to the object center makes more contribution than one that is far away.

In the course of tracking the object shape may undergo non-rigid deformation. Let $\Omega$ denote the candidate region of the object with boundary $\Gamma$, its center $\mathbf{z}_c = (x_c, y_c)$ may be computed as

$$x_c = \frac{1}{|\Omega|} \sum\limits_{\mathbf{z}_i \in \Omega} x_i$$

$$y_c = \frac{1}{|\Omega|} \sum\limits_{\mathbf{z}_i \in \Omega} y_i$$

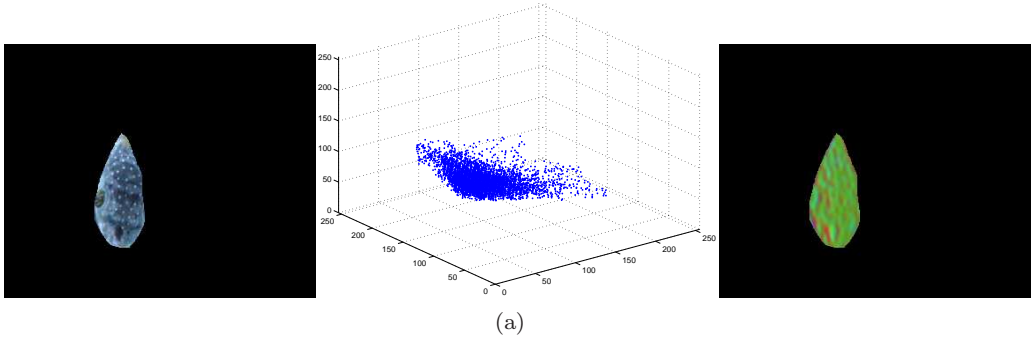

(a)

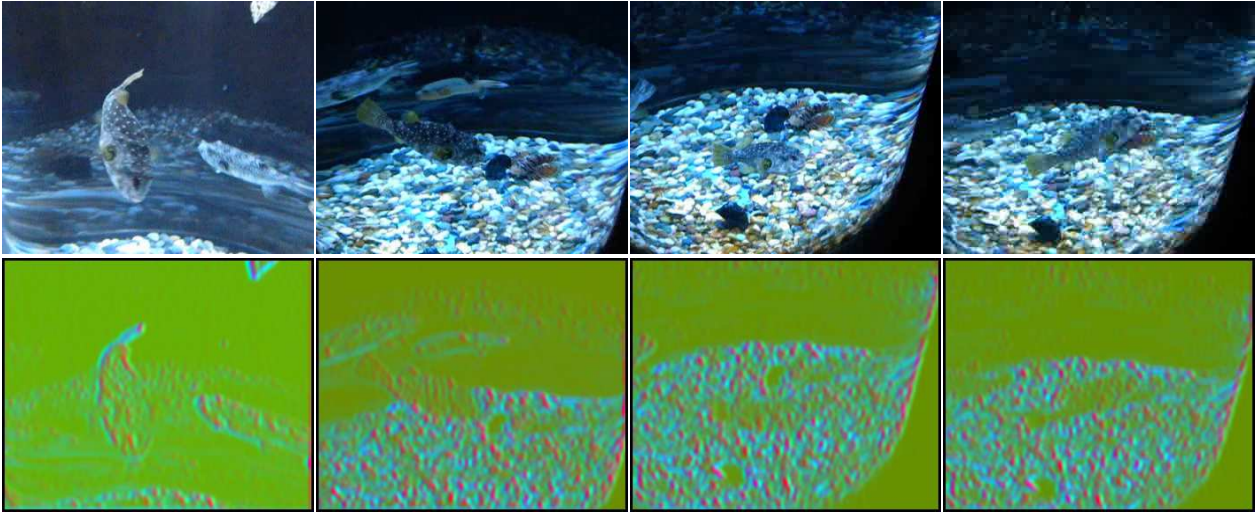

(b)

**Fig. 2** Example of the Tensor-SIFT ($K = 3$). (a) The dense SIFT features are computed in the reference image (left), and by the ALS algorithm the three-dimensional Tensor-SIFT are produced; the corresponding scatter map (middle) and tensor-SIFT image (right) are shown. (b) Given four new images (top row), the corresponding Tensor-SIFTs are computed and the tensor-SIFT images are shown (bottom row). Note that the values of the Tensor-SIFTs are scaled to $[0, 255]$ for each dimension.

where $|\Omega|$ denotes the number of points in the region $\Omega$. Then the candidate distribution $q_v(\Omega)$ for $v = 1, \cdots, V$ can be described by

$$q_v(\Omega) = \frac{1}{\sum_{\mathbf{z}_i \in \Omega} w(\mathbf{z}_i - \mathbf{z}_c)} \sum_{\mathbf{z}_i \in \Omega} w(\mathbf{z}_i - \mathbf{z}_c)\delta_v(\mathbf{I}(\mathbf{z}_i)) \quad (5)$$

Note that the symbols in the above equation have similar meanings to their counterparts in (4), and $V$ indicates that the candidate object is divided into $V$ subspaces.

The region functional that measures the distance (dissimilarity) between the reference and candidate models is defined as follows:

$$\underset{\Omega}{\operatorname{argmin}} \ f(\Omega) \quad (6)$$

$$f(\Omega) = \min_{r_{uv}} \sum_{u=1}^{U} \sum_{v=1}^{V} d_{uv} r_{uv}(\Omega) \quad (7)$$

subject to

$$\sum_{u=1}^{U} r_{uv}(\Omega) = q_v(\Omega), \ \ v = 1, \cdots, V$$

$$\sum_{v=1}^{V} r_{uv}(\Omega) = p_u, \ \ u = 1, \cdots, U$$

$$\sum_{u=1}^{U} \sum_{v=1}^{V} r_{uv}(\Omega) = \min(\sum_{u=1}^{U} p_u, \sum_{v=1}^{V}(q_u(\Omega)))$$

$$r_{uv} \geq 0, \ \ u = 1, \cdots, U, \ v = 1, \cdots, V$$

Note that in our problem, $U = V$, and the kernel densities are normalized so that the last constraint equation is redundant; 2) U=V. In the above equations the ground distance $d_{uv}$ between the clusters $u, v$ may be interpreted as the cost in transporting the unit goods from the provider $u$ to the consumer $v$. Thus, the tracking problem can be formulated as how to seek a candidate region $\Omega$ with a complex shape $\Gamma$, so that its distribution $\{q_v(\Omega), v = 1, \cdots, V\}$ has the smallest distance with the reference distribution $\{p_u, \ u = 1, \cdots, U\}$.

Optimizing the region functional (6) is nontrivial. We propose a two-phase method for the functional formulation. In the first phase, let the candidate contour be fixed. EMD is computed as the best solution to the transportation problem using the simplex algorithm. Next, using the shape derivative theory, we make a perturbation analysis of the contour around the best solution, thus deriving the corresponding PDE that governs the candidate contour evolution. Note that a similar two-phase method is used for functional minimization in [6, Sec. 2.2].

## 4.2 Simplex Algorithm for Solving EMD

Let the image region $\Omega$ be fixed. The problem described by (7) becomes the classical transportation problem, which can be solved by the simplex algorithm. Equation (7) can be expressed in matrix form as follows (the subscript $\Omega$ is omitted because $\Omega$ is fixed):

$$\min_{\mathbf{x}} \ f = \mathbf{c}^T \mathbf{x}$$
$$\text{subject to } \ \mathbf{A}\mathbf{x} = \mathbf{b}$$
$$\mathbf{x} \geq \mathbf{0} \qquad (8)$$

Here $\mathbf{c}$, $\mathbf{x}$ are both vectors of $UV$, $\mathbf{b}$ is a vector of $U + V + 1$, and $\mathbf{A}$ is a matrix of $U + V + 1$ by $UV$;

they have the following forms, respectively:

$$\mathbf{c} = \begin{bmatrix} d_{11} \ d_{12} \ \cdots \ d_{1V} \ \cdots \ d_{U1} \ d_{U2} \ \cdots \ d_{UV} \end{bmatrix}$$

$$\mathbf{x} = \begin{bmatrix} r_{11} \ r_{12} \ \cdots \ r_{1V} \ \cdots \ r_{U1} \ r_{U2} \ \cdots \ r_{UV} \end{bmatrix}$$

$$\mathbf{b} = \begin{bmatrix} q_1 \ q_2 \ \cdots \ q_V \ \cdots \ p_1 \ p_2 \ \cdots \ p_U \ 1 \end{bmatrix}$$

$$\mathbf{A} = \begin{bmatrix} e_{11} & 0 & \cdots & 0 & \cdots & e_{U1} & 0 & \cdots & 0 \\ & & & & \vdots & & & & \\ 0 & 0 & \cdots & e_{1V} & \cdots & 0 & 0 & \cdots & e_{UV} \\ e_{11} & e_{12} & \cdots & e_{1V} & \cdots & 0 & 0 & \cdots & 0 \\ & & & & \vdots & & & & \\ 0 & 0 & \cdots & 0 & \cdots & e_{U1} & e_{U2} & \cdots & e_{UV} \\ e_{11} & e_{12} & \cdots & e_{1V} & \cdots & e_{U1} & e_{U2} & \cdots & e_{UV} \end{bmatrix}$$

In the matrix $\mathbf{A}$ all the elements $e_{ij} = 1$, $i = 1, \cdots, U$, $j = 1, \cdots, V$.

Let $\mathbf{x} = [\mathbf{x_B} \ \mathbf{x_N}]$ be an initial feasible solution, where $\mathbf{x_B}$ and $\mathbf{x_N}$ denote the vectors of basic variables and non-basic variables, respectively. Accordingly, the vector $\mathbf{c}$ can be written as $\mathbf{c} = [\mathbf{c_B} \ \mathbf{c_N}]$. In a similar manner, the matrix $\mathbf{A}$ can be written as $\mathbf{A} = [\mathbf{B} \ \mathbf{N}]$, where $\mathbf{B}$, comprising columns in $\mathbf{A}$ that correspond to the basic variables, is a basis of the Euclidean space $\mathbb{R}^m (m = U + V + 1)$, and $\mathbf{N}$ consists of columns in $\mathbf{A}$ corresponding to the non-basic variables.

Starting from the initial feasible solution, the simplex algorithm proceeds through an iteration process. This process can be explained in tableau form. Table 1 illustrates the initial simplex tableau, where $\mathbf{0}$ and $\mathbf{I}$ denote the zero vector and unitary vector, respectively. Each iteration seeks an improved feasible solution that decreases the value of the function $f$. Consider one iteration. The most negative component of $\mathbf{r_N^T} = \mathbf{c_B^T} \mathbf{B}^{-1} \mathbf{N} - \mathbf{c_N^T} \leq \mathbf{0}$ is determined, and the corresponding column in $\mathbf{N}$, denoted by $\mathbf{N}_j$, is selected as the vector to enter the basis matrix $\mathbf{B}$. Next, $\mathbf{y} = \mathbf{B}^{-1} \mathbf{N}_j$ and $\tilde{\mathbf{b}} = \mathbf{B}^{-1} \mathbf{b}$ are computed and then the pivoting row index $i$ is decided for which $i = \max_k \{ y_k / \tilde{b}_k, y_k > 0 \}$, where $y_k$ and $\tilde{b}_k$ are components of $\mathbf{y}$ and $\tilde{\mathbf{b}}$, respectively. After the pivot element is determined, we can perform Gaussian elimination to update $\mathbf{B}^{-1}$. The column vector in $\mathbf{B}$ that corresponds to the $i$th basic variable of $\mathbf{x_B}$ is selected to leave the basis matrix $\mathbf{B}$.

**Table 1** The simplex tableau

| | $f$ | $\mathbf{x_B^T}$ | $\mathbf{x_N^T}$ | |
|---|---|---|---|---|
| $\mathbf{x_B}$ | 0 | $\mathbf{I}$ | $\mathbf{B^{-1}N}$ | $\mathbf{B^{-1}b}$ |
| $f$ | 1 | $-\mathbf{c_B^T}$ | $\mathbf{c_B^T B^{-1}N - c_N^T}$ | $\mathbf{c_B^T B^{-1}b}$ |

Once $\mathbf{r_N^T} > \mathbf{0}$ is identified during the iterations, the best solution to the transportation problem is obtained, and the iteration should be terminated. The EMD between the candidate and reference models is given by

$$f(\Omega) = \mathbf{c_B^T B^{-1} b} \qquad (9)$$
$$= \sum_{v=1}^{V} l_v q_v(\Omega) + \sum_{u=1}^{U} h_u p_u + c$$

where

$$l_v = \sum_{i=1}^{U+V+1} \mathbf{c_b}(i) \mathbf{B}^{-1}(i,v), \qquad v = 1, \cdots, V$$
$$h_u = \sum_{i=1}^{U+V+1} \mathbf{c_b}(i) \mathbf{B}^{-1}(i,u+V), \quad u = 1, \cdots, U$$
$$c = \sum_{i=1}^{U+V+1} \mathbf{c_b}(i) \mathbf{B}^{-1}(i,U+V+1)$$

Here $\mathbf{c_b}(i)$ denotes the $i$th component in the vector, and $\mathbf{B}^{-1}(i,v)$ the entry of row $i$ and column $v$ in the matrix.

The best solution always exists for the balanced transportation problem [30]. Note that in such case, the matrix $\mathbf{A}$ is not full-rank ($U + V - 1$ in our case). We introduce artificial variables and obtain the initial feasible solution according to the method described in Luenberger and Ye [30, pp. 50–54].

## 4.3 PDE Derivation with Theory of Shape Derivative

After the first phase is completed, we can perform a perturbation analysis around the best solution to the transportation problem. That is, let $\Omega$ be a variable and solve the functional (9). The second and third terms on the right-hand side of (9) are irrelevant to $\Omega$ and therefore can be be discarded. Hence, the functional to be optimized becomes

$$f(\Omega) = \sum_{v=1}^{V} l_v q_v(\Omega) \qquad (10)$$

Deriving the PDE associated with this functional is not straightforward. Here the theory of shape derivative is used.

### 4.3.1 Shape Derivative

Given an initial domain (region) $\Omega$, the set of all the possible deformations may not be a vector space. The theory of shape derivative introduces a family of deformations (transformations) $\{\Omega_\tau, \ \tau > 0\}$ such that the perturbation of $\Omega$ is possible with respect to $\tau$.

Suppose that the perturbation of a point $\mathbf{z} \in \Omega$ is governed by the differential equation

$$\frac{d\mathbf{z}(\tau)}{dt} = \mathbf{V}(\mathbf{z}(\tau)), \ \tau > 0, \ \mathbf{z}(0) = \mathbf{z} \qquad (11)$$

where $\mathbf{V}$ is a vector field. We can define the transformations of a point and the region as follows:

$$T(\tau, \mathbf{z}) \triangleq \mathbf{z}(\tau), \ \tau > 0, \ \mathbf{z}(0) = \mathbf{z} \qquad (12)$$
$$\Omega_\tau \triangleq T_\tau(\Omega) = \{T(\tau, \mathbf{z}), \ \mathbf{z} \in \Omega\}$$

The region functional $J(\Omega) = \int_\Omega \phi(\Omega)$ is said to have a *Eulerian derivative* or a *shape derivative*, at $\Omega$ in the direction of the vector field $\mathbf{V}$ if the following limit exists and is finite

$$dJ(\Omega; \mathbf{V}) = \lim_{\tau \to 0} \frac{J(\Omega_\tau) - J(\Omega)}{\tau} \qquad (13)$$

The functional $J(\Omega)$ is shape differentiable at $\Omega$ if the Eulerian derivative $dJ(\Omega; \mathbf{V})$ exists for all directions $\mathbf{V}$ and the mapping $\mathbf{V} \to J(\Omega; \mathbf{V})$ is linear and continuous.

The *material derivative* of a function $\phi(\Omega)$ at $\Omega$ in the direction of the vector field $\mathbf{V}$ is defined as

$$\dot{\phi}(\Omega; \mathbf{V}) = \lim_{\tau \to 0} \frac{\phi(\Omega_\tau) \circ T(\tau, \mathbf{z}) - \phi(\Omega)}{\tau} \qquad (14)$$

And the *shape derivative* of a function $\phi(\Omega)$ at $\Omega$ in the direction of the vector field $\mathbf{V}$ is defined as

$$\phi'(\Omega; \mathbf{V}) = \lim_{\tau \to 0} \frac{\phi(\Omega_\tau) \circ T(0, \mathbf{z}) - \phi(\Omega)}{\tau} \qquad (15)$$

The following theorem establishes the relation between the Eulerian derivative and the shape derivatives:

**Theorem 1** (Theorem on the Eulerian and shape derivatives) *If the material and shape derivatives of a function $\phi(\Omega)$ exist, then the functional $J(\Omega) = \int_\Omega \phi(\Omega)$ is shape differentiable and the following equation holds*

$$dJ(\Omega; \mathbf{V}) = \int_\Omega \phi'(\Omega; \mathbf{V})d\mathbf{z} - \int_\Gamma \phi(\Omega)\langle \mathbf{V}, \mathbf{N}\rangle ds \quad (16)$$

where $\langle \mathbf{V}, \mathbf{N}\rangle$ denotes the inner product between $\mathbf{V}$ and $\mathbf{N}$.

For the complete and strict definitions of and theories about shape derivative, refer to [50] and [12]. A brief yet insightful introduction about shape derivative is given in Aubert et al. [1].

### 4.3.2 Derivation of the PDE through Shape Derivative

Our derivation of the PDE to optimize the region functional builds upon theorem 1. First we write the candidate distribution in a continuous form:

$$q_v(\Omega) = \frac{1}{\int_\Omega w(\mathbf{z} - \mathbf{z}_c)d\mathbf{z}} \int_\Omega w(\mathbf{z} - \mathbf{z}_c)\delta_v(\mathbf{I}(\mathbf{z}))d\mathbf{z} \quad (17)$$

where the Gaussian kernel $w(\mathbf{z} - \mathbf{z}_c) = \exp(-\|\mathbf{z} - \mathbf{z}_c\|^2/(2\sigma^2))$ is adopted, and the object center $\mathbf{z}_c = [x_c \ y_c]^T$ takes the form

$$\begin{cases} x_c = \frac{1}{\int_\Omega d\mathbf{z}}\int_\Omega x d\mathbf{z} \\ y_c = \frac{1}{\int_\Omega d\mathbf{z}}\int_\Omega y d\mathbf{z} \end{cases}$$

To facilitate derivation, we write $x_c, y_c$ in the form

$$x_c = \frac{G_1(\Omega)}{G_0(\Omega)}, \quad G_1(\Omega) = \int_\Omega x d\mathbf{z}, \quad G_0(\Omega) = \int_\Omega d\mathbf{z}$$

$$y_c = \frac{G_2(\Omega)}{G_0(\Omega)}, \quad G_2(\Omega) = \int_\Omega y d\mathbf{z} \quad (18)$$

We also re-express the candidate distribution as follows:

$$q_v(\Omega) = \frac{K_1(\Omega)}{K_2(\Omega)}$$

$$K_1(\Omega) = \int_\Omega L_1(\Omega)d\mathbf{z}, \quad L_1(\Omega) = w(\mathbf{z} - \mathbf{z}_c)\delta_v(\mathbf{I}(\mathbf{z}))$$

$$K_2(\Omega) = \int_\Omega L_2(\Omega))d\mathbf{z}, \quad L_2(\Omega) = w(\mathbf{z} - \mathbf{z}_c)) \quad (19)$$

The Eulerian derivative of the region functional (10) is given by the formula

$$df(\Omega; \mathbf{V}) = \sum_{v=1}^V l_v dq_v(\Omega; \mathbf{V}) \quad (20)$$

$$= \sum_{v=1}^V l_v \left( \frac{\partial q_v}{\partial K_1}dK_1(\Omega; \mathbf{V}) + \frac{\partial q_v}{\partial K_2}dK_2(\Omega; \mathbf{V}) \right)$$

where $\partial q_v/\partial K_1$ is the partial derivative of $q_v$ with respect to $K_1$ and $\partial q_v/\partial K_2$ has a similar meaning.

Now consider the $dK_1(\Omega; \mathbf{V})$ in (20). From theorem 1 we have

$$dK_1(\Omega; \mathbf{V}) = \int_\Omega L_1'(\Omega; \mathbf{V})d\mathbf{z} - \int_\Gamma L_1(\Omega)\langle \mathbf{V}, \mathbf{N}\rangle ds \quad (21)$$

From (18) and (19), $L_1$ is a function of $G_0$, $G_1$ and $G_2$ which are all functionals of $\Omega$. Hence, using the chain rule of the shape derivative we get

$$L_1'(\Omega; \mathbf{V}) = \frac{\partial L_1}{\partial G_0}dG_0(\Omega; \mathbf{V}) + \frac{\partial L_1}{\partial G_1}dG_1(\Omega; \mathbf{V})$$

$$+ \frac{\partial L_1}{\partial G_2}dG_2(\Omega; \mathbf{V}) \quad (22)$$

The shape derivatives of the integrands in $G_0$, $G_1$, and $G_0$ are all zero because they are independent of $\Omega$. Hence, application of theorem 1 to $G_0$, $G_1$, and $G_0$, respectively, leads to the following equations:

$$dG_0(\Omega; \mathbf{V}) = -\int_\Gamma \langle \mathbf{V}, \mathbf{N}\rangle ds$$

$$dG_1(\Omega; \mathbf{V}) = -\int_\Gamma x\langle \mathbf{V}, \mathbf{N}\rangle ds$$

$$dG_2(\Omega; \mathbf{V}) = -\int_\Gamma y\langle \mathbf{V}, \mathbf{N}\rangle ds \quad (23)$$

Note that the partial derivatives of $L_1$ with respect to $G_0$, $G_1$ and $G_2$ are in the following forms:

$$\frac{\partial L_1}{\partial G_0} = -\frac{w'(\mathbf{z} - \mathbf{z}_c)}{\sigma^2 G_0}((x - x_c)x_c + (y - y_c)y_c)\delta_v(\mathbf{I}(\mathbf{z}))$$

$$\frac{\partial L_1}{\partial G_1} = \frac{w'(\mathbf{z} - \mathbf{z}_c)}{\sigma^2 G_0}(x - x_c)\delta_v(\mathbf{I}(\mathbf{z}))$$

$$\frac{\partial L_1}{\partial G_2} = \frac{w'(\mathbf{z} - \mathbf{z}_c)}{\sigma^2 G_0}(y - y_c)\delta_v(\mathbf{I}(\mathbf{z})) \tag{24}$$

Combining (21), (22), (23), and (24) and after some manipulation, we obtain

$$dK_1(\Omega; \mathbf{V}) =$$
$$\int_\Gamma \left( -\frac{1}{\sigma^2 G_0} \int_\Omega (\tilde{\mathbf{z}} - \mathbf{z}_c)^T (\mathbf{z} - \mathbf{z}_c) w'(\tilde{\mathbf{z}} - \mathbf{z}_c) \right.$$
$$\left. \delta_v(\mathbf{I}(\tilde{\mathbf{z}}))d\tilde{\mathbf{z}} - w(\mathbf{z} - \mathbf{z}_c)\delta_v(\mathbf{I}(\mathbf{z})) \right) \langle \mathbf{V}, \mathbf{N} \rangle ds \tag{25}$$

where $\tilde{\mathbf{z}} = [\tilde{x} \ \tilde{y}]^T$ is a point in the region $\Omega$, and $\mathbf{z} = [x \ y]^T$ is a point on the boundary $\Gamma$.

Likewise, we can derive

$$dK_2(\Omega; \mathbf{V}) =$$
$$\int_\Gamma \left( -\frac{1}{\sigma^2 G_0} \int_\Omega (\tilde{\mathbf{z}} - \mathbf{z}_c)^T (\mathbf{z} - \mathbf{z}_c) w'(\tilde{\mathbf{z}} - \mathbf{z}_c) d\tilde{\mathbf{z}} \right.$$
$$\left. - w(\mathbf{z} - \mathbf{z}_c) \right) \langle \mathbf{V}, \mathbf{N} \rangle ds \tag{26}$$

Note that $\partial q_v / \partial K_1 = 1/K_2 and \partial q_v / \partial K_2 = q_v/K_2$. Substituting (25) and (26) into (20), we have the complete expression of the Eulerian derivative of the region functional $f(\Omega)$:

$$df(\Omega; \mathbf{V}) = \int_\Gamma F \langle \mathbf{V}, \mathbf{N} \rangle ds \tag{27}$$

where $F$ is of the following form

$$F = -\frac{1}{\sigma^2 G_0 K_2} \int_\Omega (\tilde{\mathbf{z}} - \mathbf{z}_c)^T (\mathbf{z} - \mathbf{z}_c) w'(\tilde{\mathbf{z}} - \mathbf{z}_c)$$
$$\times \left( \sum_{v=1}^V l_v (\delta_v(\mathbf{I}(\tilde{\mathbf{z}})) - q_v) \right) d\tilde{\mathbf{z}}$$
$$- \frac{1}{K_2} w(\mathbf{z} - \mathbf{z}_c) \left( \sum_{v=1}^V l_v (\delta_v(\mathbf{I}(\mathbf{z})) - q_v) \right) \tag{28}$$

From the above shape derivatives, we obtain the PDE associated with the region functional (10) governing the contour evolution

$$\frac{\partial \Gamma}{\partial \tau} = F \mathbf{N} \tag{29}$$

The preceding PDE is obtained when the normal kernel is employed. If other kernels are used, we can derive the corresponding PDEs in a similar manner. Table 2 lists $F$ with respect to the normal kernel, Epanechnikov kernel and uniform kernel, respectively. The first term in $F$, when the non-uniform kernels are used, is due to the spatial weight introduced, reflecting the combined effect of one point $\mathbf{z}$ on the contour and the points within the region. Our experiments show that the normal kernel performs well and thus it is used in this paper. The parameter $\sigma$ is selected as half of the radius of the minimal enclosing circle of the contour.

Similar to Aubert et al. [1], we also impose a smooth constraint on the contour by adding a second term $\alpha \int_\Gamma ds$ in Eq. (6), where $\alpha$ is a constant. The minimization of this curve length term gives rise to a curvature term in the PDE [5]. Thus, our final PDE has the following form [1]:

$$\frac{\partial \Gamma}{\partial \tau} = (F + \alpha \kappa)\mathbf{N} \tag{30}$$

where $\kappa$ is the curvature of the curve $\Gamma$.

### 4.3.3 Solution to PDE through the Level Set Method

The idea of the level set method is that at any time the contour $\Gamma$ is implicitly represented by a zero level set of a higher-dimensional function $\phi$, i.e.,

$$\Gamma(\mathbf{z}, \tau) = \{\mathbf{z} : \phi(\mathbf{z}, \tau) = 0\}, \ \text{ given } \Gamma(\mathbf{z}, 0)$$

We define $\phi(\mathbf{z}, \tau) < 0$ in the interior region and $\phi(\mathbf{z}, \tau) > 0$ in the exterior region. Taking the partial derivative of $\phi$ with respect to $\tau$ and using the chain rule, we have

$$\frac{\partial \phi}{\partial \tau} + \nabla \phi \cdot \frac{\partial \Gamma}{\partial \tau} = 0 \tag{31}$$

where $\nabla \phi = [\phi_x \ \phi_y]^T$ is the gradient of $\phi$ with respect to $\mathbf{z}$. Note that $\mathbf{N} = -\nabla \phi / \|\nabla \phi\|$ and $\kappa = -\text{div} (\nabla \phi / \|\nabla \phi\|)$, where div denotes the divergence. Substituting (30) into (31), we have the following PDE:

$$\phi_\tau = \left( F + \alpha \triangle \left( \frac{\nabla \phi}{\|\nabla \phi\|} \right) \right) \|\nabla \phi\| \tag{32}$$

**Table 2** The function $F$ with respect to various kernel

| Kernel | The function $F$ |
|---|---|
| Normal kernel<br>$w_N(\tilde{\mathbf{z}}) = \exp(-\frac{\|\tilde{\mathbf{z}} - \mathbf{z}_c\|^2}{2\sigma^2})$ $\tilde{\mathbf{z}} \in \Omega$ | $-\frac{1}{\sigma^2 G_0 K_2} \int_\Omega (\tilde{\mathbf{z}} - \mathbf{z}_c)^T (\mathbf{z} - \mathbf{z}_c) w_N'(\tilde{\mathbf{z}} - \mathbf{z}_c) \left( \sum_{v=1}^V l_v(\delta_v(\mathbf{I}(\tilde{\mathbf{z}})) - q_v) \right) d\tilde{\mathbf{z}}$<br>$\quad -\frac{1}{K_2} w_N (\mathbf{z} - \mathbf{z}_c) \left( \sum_{v=1}^V l_v(\delta_v(\mathbf{I}(\mathbf{z})) - q_v) \right)$ |
| Epanechnikov Kernel<br>$w_E(\tilde{\mathbf{z}}) = \begin{cases} 1 - \|\frac{\tilde{\mathbf{z}} - \mathbf{z}_c}{h}\|^2 & \tilde{\mathbf{z}} \in \Omega, \ \|\frac{\tilde{\mathbf{z}} - \mathbf{z}_c}{h}\| \le 1 \\ 0 & \text{otherwise} \end{cases}$ | $-\frac{2}{h^2 G_0 K_2} \int_\Omega (\tilde{\mathbf{z}} - \mathbf{z}_c)^T (\mathbf{z} - \mathbf{z}_c) \left( \sum_{v=1}^V l_v(\delta_v(\mathbf{I}(\tilde{\mathbf{z}})) - q_v) \right) d\tilde{\mathbf{z}}$<br>$\quad -\frac{1}{K_2} w_E(\mathbf{z} - \mathbf{z}_c) \left( \sum_{v=1}^V l_v(\delta_v(\mathbf{I}(\mathbf{z})) - q_v) \right)$ |
| Uniform Kernel<br>$w_U = \begin{cases} 1 & \tilde{\mathbf{z}} \in \Omega \\ 0 & \text{otherwise} \end{cases}$ | $-\frac{1}{G_0} \left( \sum_{v=1}^V l_v(\delta_v(\mathbf{I}(\mathbf{z})) - q_v) \right)$ |

Applying to (32) the first-order forward time difference scheme, the upwind scheme for the first-order discretization of $F$, and the first and second-order central difference schemes for the curvature $\kappa$, we have the following discrete equation:

$$\phi_{i,j}^{\tau+1} = \phi_{i,j}^\tau + \Delta\tau \left( \max(F_{i,j}^\tau, 0)^2 \nabla^{\tau+} \right.$$
$$\left. + \min(F_{i,j}^\tau, 0)^2 \nabla^{\tau-} + \alpha \kappa_{i,j}^\tau \right) \quad (33)$$

where $\phi_{i,j}^\tau$ denotes the function value on the grid point $(i, j)$ at iteration step $\tau$, $\Delta\tau$ denotes the discrete time interval, and $\nabla^{\tau+}$, $\nabla^{\tau-}$ and $\kappa_{i,j}^\tau$ take the following forms, respectively:

$$\nabla^{\tau+} = \left( \max(D_{i,j}^{-x}, 0)^2 + \min(D_{i,j}^{+x}, 0)^2 \right.$$
$$\left. + \max(D_{i,j}^{-y}, 0)^2 + \min(D_{i,j}^{+y}, 0)^2 \right)^{1/2}$$
$$\nabla^{\tau+} = \left( \max(D_{i,j}^{+x}, 0)^2 + \min(D_{i,j}^{-x}, 0)^2 \right.$$
$$\left. + \max(D_{i,j}^{+y}, 0)^2 + \min(D_{i,j}^{-y}, 0)^2 \right)^{1/2}$$
$$\kappa_{i,j}^\tau = \left( D_{i,j}^{+x+x}(D_{i,j}^{0y})^2 - 2D_{i,j}^{0x}D_{i,j}^{0y}D_{i,j}^{xy} \right.$$
$$\left. + D_{i,j}^{+y+y}(D_{i,j}^{0x})^2 \right) / \left( (D_{i,j}^{0x})^2 + (D_{i,j}^{0y})^2 \right)$$

Here a short-hand notation is used where the operator $D^{-x}\phi_{i,j}^\tau$ is written as $D_{i,j}^{-x}$, etc. The operators involved are defined as follows:

$$D_{i,j}^{-x} = \phi_{i,j}^\tau - \phi_{i-1,j}^\tau$$
$$D_{i,j}^{+x} = \phi_{i+1,j}^\tau - \phi_{i,j}^\tau$$
$$D_{i,j}^{0x} = (\phi_{i+1,j}^\tau - \phi_{i-1,j}^\tau)/2$$
$$D_{i,j}^{+x+x} = (\phi_{i+1,j}^\tau - 2\phi_{i,j}^\tau + \phi_{i-1,j}^\tau)/2$$
$$D_{i,j}^{+x+y} = (\phi_{i+1,j+1}^\tau - \phi_{i+1,j-1}^\tau - \phi_{i-1,j+1}^\tau$$
$$+ \phi_{i-1,j-1}^\tau)/2$$

The operators $D_{i,j}^{-y}$, $D_{i,j}^{+y}$, $D_{i,j}^{0y}$ and $D_{i,j}^{+y+y}$ have similar forms. At each iteration step $\tau$, numerical stability can be guaranteed by the following CFL condition [36]

$$\Delta\tau \max_{i,j} \left\{ F_{i,j}^\tau \frac{\text{abs}(D_{i,j}^{+x}) + \text{abs}(D_{i,j}^{+y})}{((D_{i,j}^{0x})^2 + (D_{i,j}^{0y})^2)^{1/2}} + 4\alpha \right\} < 1 \quad (34)$$

The implicit function $\phi$ can be arbitrary and is usually chosen as a signed distance function $\phi(\mathbf{z}, \tau) = d(\mathbf{z}, \tau)$. As suggested [36, ch.7], in the level set algorithm, the re-initialization technique is required periodically so that $\phi(\mathbf{z}, \tau)$ still remains a valid signed distance function during the evolution process. We use the fast marching method proposed by Sethian [49, ch.8] for re-initialization and the re-initialization frequency of 50 suffices for all of our experiments.

## 5 Integrated Contour Tracking Algorithm

This section describes the integrated tracking algorithm (section 5.1), followed by the discussion of the implementation detail (section 5.2) and the computational cost analysis (section 5.3).

### 5.1 Integrated Tracking Algorithm

The level set algorithm is computationally expensive. In tracking applications, because we are only concerned with the object contour, i.e., the zero level set, we update in each iteration only the values of the level set function in a narrow band in the neighboring region around the current zero level set. Re-initialization of the level set function using the fast marching algorithm is also performed within the narrow band.

In contour tracking algorithms, it is not straightforward to exploit temporal continuity. Rathi et al. [46] proposed a method in the particle filter framework. However, the high computational cost of the particle filter plus that of the contour evolution using the level set algorithm may be a very heavy load. We introduce a scheme that uses the mean shift algorithm [8] for exploiting temporal continuity, which provides inter-frame contour initialization. In frame $t$, before PDE iteration starts, we first use an ellipse fitting method [14] to fit the contour $\Omega^*(t-1)$, i.e., the tracking result in frame $t-1$. Next, we perform the mean shift tracking algorithm until its convergence and then enlarge the major and minor radii of the resulting ellipse by twenty percent, respectively. Let $(x_0, y_0, a, b, \theta)$ be the parameters of the resulting ellipse, where $(x_0, y_0)$ denotes the ellipse center, $a, b$ denote the radius along $x-$ and $y-$ axes, and $\theta$ indicates its orientation. We construct the initial level set function through the following formula

$$\phi_{i,j}^{\tau=0}(t) = \frac{((i-x_0)\cos\theta + (j-y_0)\sin\theta)^2}{a^2}$$
$$+ \frac{((i-x_0)\sin\theta - (j-y_0)\cos\theta)^2}{b^2} - 1$$

This scheme gives a kind of contour measurement, taking advantage of temporal continuity and thus providing a more accurate initial contour that helps avoid possible local minimum in gradient descent iteration.

There is no general stopping criterion in the level set algorithm and the fixed number of iterations is often used. As EMD is robust in measuring the distance between the target and candidate models, we introduce a stopping criterion that ends the iteration if the EMD does not decrease. Practically, we keep the latest 20 values of EMD to which a straight line is fitted by the least square method, and terminate the iteration process if the line slope is larger than or equal to zero. Due to temporal continuity, the object size change may not be large. Thus, we also terminate the iteration if the area difference between $\Omega^*(t-1)$ an $\Omega_\tau(t)$ is larger than ten percent.

Given the primary considerations above, we now describe the complete tracking algorithm in Algorithm 1.

### 5.2 Implementation Detail

This section discusses the implementation detail in the tracking algorithm.

*Selection of $K$ in tensor decomposition* The object size is usually small in a tracking task; thus, we select $8\times8$ sampling window in the SIFT feature computation. In the tensor decomposition problem, what $K$ best fits the original tensor is the problem of the best rank-$k$ approximation that is under investigation [23], which is beyond the scope of this paper. Intuitively, larger $K$ may better explain the original tensor, which, however, also means the larger dimensionality of tensor-SIFT. In this paper we demonstrate conceptually our idea by simply setting $K = 3$.

*Parameters setting in inter-frame initialization* Inter-frame initialization mainly concerns the mean shift iteration. Following [9], the histogram of $16\times16\times16$ bins and the maximum iteration number 10 are used for all experiments. Note that in this case the ellipse describing the object may not be upright, and we use the ellipse filling algorithm for efficient histogram computation.

*Computations of Signatures and EMD* In accordance with [48], the ground distance $d_{uv}$ is defined as

$$d_{uv} = 1 - e^{-\beta\|\mathbf{h}_u^* - \mathbf{h}_v\|} \quad \text{with} \quad \beta = \|[\zeta_1 \ \dots \ \zeta_K]\|$$

where $\zeta_k$ denotes the standard deviation of component $k$, learned from all reference feature vectors. $d_{uv}$ saturates to 1 for a large distance between two clusters, avoiding the side effect that few large distances brings on the overall distance. The efficient

**1** Provided the reference image of the object, compute its dense SIFT features; make tensor decomposition using ALS algorithm in section 3.2, achieving the reference tensor-SIFT image and the tensor basis matrices **R**, **S** and **T**;

**2** Produce the reference signature $(\mathbf{h}_u^*, p_u)$ for $u = 1, \cdots, U$ with tensor-SIFT features;

**3** Initialize manually the contour in the first frame $\Omega_0(t = 1)$;

**4 for** $t = 2, 3, \ldots$ **do**

    **5**    For the image $\mathbf{D}(t, \mathbf{z})$, compute its SIFT features and then the corresponding Tensor-SIFT image $\mathbf{I}(t, \mathbf{z})$ by projection from (3);

    **6**    Construct the initial level set function $\phi_{i,j}^{\tau=0}(t)$ from the tracking result $\Omega^*(t-1)$, using a scheme of inter-frame initialization (described in the text);

    **7**    $\tau = 0$;

    **8**    **while** *Stoping criterion is not true* **do**

        **9**    Produce the candidate signature $(\mathbf{h}_v, q_v)$ for $v = 1, \cdots, V$ within $\Omega_\tau(t) = \{(i, j) | \phi_{i,j}^\tau(t) < 0\}$;

        **10**   Calculate the EMD between the reference and candidate signatures by the simplex algorithm in section 4.2;

        **11**   Extract a narrow band $S_{narr}$ in the neighboring region around the zero level set $\Gamma_\tau(t) = \{(i, j) | \phi_{i,j}^\tau(t) = 0\}$;

        **12**   **for** *each grid point* $(i, j) \in S_{narr}$ **do**

            **13**   Compute its force $F_{i,j}^\tau$;

            **14**   Compute its curvature $\kappa_{i,j}^\tau$

        **15**   **end**

        **16**   Calculate the discrete time interval $\triangle\tau$ from (34);

        **17**   **for** *each grid point* $(i, j) \in S$ **do**

            **18**   Update the level set function $\phi_{i,j}(\tau)$ from (33)

        **19**   **end**

        **20**   $\tau = \tau + 1$;

        **21**   Re-initialize the level set function every 50 iterations using the fast marching algorithm;

    **22**    **end**

    **23**    Assign tracking result $\Omega^*(t) = \Omega_\tau(t)$.

**24 end**

**Algorithm 1:** Tracking algorithm

K-D tree-based clustering method [48] is employed for color space partition.

*Selection of $\alpha$ in PDE* The term of the curve length functional leads to a curvature term in PDE (30), which makes the contour to have a tendency to shrink and actually imposes a smoothness constraint on the contour. The value of $\alpha$ reflects the effect of this term and is empirically set to 0.0002 in all the experiments.

## 5.3 Computational Cost Analysis

In the preceding tracking algorithm, the procedures that dominate the computational cost are (a) computations of SIFT features and Tensor-SIFT image; (b) inter-frame initialization; and (c) functional optimization.

As dense SIFT features are required, we perform the computation for each grid point in the image. With the SIFT features at hand, the computation of the Tensor-SIFT image is accomplished by matrix multiplication, because as shown in (3), the Moore-Penrose pseudo-inverse can be obtained beforehand once the tensor basis matrices are produced. Thus, the cost of this procedure may be written in the form $C_{tens} = N_{sift}(c_{sift} + c_{ten\_dec})$, where $N_{sift}$ is the number of grid points, $c_{sift}$ is the cost for computing one SIFT feature, and $c_{ten\_dec} = 128Kc_{ops}$ denotes the cost for computing Tensor-SIFT by projection, with $c_{ops}$ indicting the cost of long operation (multiplication or division). In our experiments, the computations of SIFT features and Tensor-SIFT images are accomplished off-line. Note that these computations are well-suited for parallel implementation [55].

The inter-frame initialization mainly involves ellipse fitting, mean shift algorithm, and the initial level set function construction. The ellipse fitting procedure is principally concerned with the generalized symmetric eigenvalue problem [14], whose computational cost is approximately multiples of $n^3$ long operations [40, Chap. 15], where $n$ is the matrix size. In this problem $n = 6$, thus, its computational cost is small compared with the other procedures and can thus be discarded. The cost of the mean shift algorithm can be represented as $N_{ms}c_{ms}$, where $N_{ms}$ denotes the mean shift iteration number, and $c_{ms}$ denotes the cost in one iteration. For details, refer to Comaniciu et al. [9]. The cost of the initial level set function construction may be approximated as $N_{cons} \approx 8Ac_{ops}$, where $A$ denotes the image area.

The procedure of functional optimization mainly involves two parts: computing the EMD and the level set function update. The EMD computation consists in the simplex algorithm, whose computational complexity is theoretically exponential in problem size $m$. However, in cases where $m$ is not large, the iteration number $N_{emd}$ for the simplex algorithm to converge from a feasible solution to the best one may be a small multiple of $m$: $N_{emd} = 2m \sim 3m$ [30, chap. 5]. Thus the cost of computing the EMD may be written as $C_{emd} = N_{emd}(c_{pivo} + c_{gaus})$, where $c_{pivo}$ denotes the cost to obtain the pivot element, and $c_{gaus}$ denotes the cost for performing the Gaussian elimination for the linear system of constraint equations.

The procedure of the level set function update mainly involves the evaluation of force $F$ and curvature $\kappa$. As we re-initialize the level set function by fast marching algorithm only once every 50 iterations, the cost induced by this process is thus omitted. Denote by $c_{force}$ and $c_\kappa$ the cost for the force and curvature at one grid point. The cost for the level set update is represented by $C_{pde} = |S_{narr}|(c_{force} + c_\kappa)$, where $|S_{narr}|$ denotes the grid point number in the narrow band $S_{narr}$. The dominant operation in $C_{pde}$ is the computation of the force $F$. From (28), the cost for evaluation of $F$ has the form $c_{force} = |\Omega|c_f$, where $|\Omega|$ denotes the number of grid points in the candidate region $\Omega$, $c_f$ denotes the cost involved by one grid point, mainly including the cost of dozens of multiplications and one exponential function evaluation.

In summary, the overall computational cost $C_{tota}$ of the tracking algorithm may be described as

$$
\begin{aligned}
C_{tota} &\approx C_{inter} + C_{tens} + C_{optm} \\
&\approx C_{inter} + C_{tens} + N_{optm}(C_{emd} + C_{pde}) \\
&\approx N_{ms}c_{ms} + 8Ac_{ops} \\
&+ N_{sift}(c_{sift} + 128Kc_{ops}) \\
&+ N_{optm}\big(N_{emd}(c_{pivo} + c_{gaus}) \\
&+ |S_{narr}|(|\Omega|c_f + c_\kappa)\big)
\end{aligned}
\tag{35}
$$

Here, $C_{optm}$ denotes the cost of the functional optimization, and $N_{optm}$ denotes the number of iterations involved. $N_{optm}$ is application dependent and is averaged by several hundreds in our experiments.

## 6 Experiments and Discussion

We assess the performance of three trackers, Color_EMD, PCA_SIFT_EMD, and Tensor_SIFT_EMD, which are all based on EMD but use varying features, as their respective names indicate. We also compare our trackers with the tracker based on the Bhattacharyya coefficient that use color histogram [16] (henceforth Color_Bhattacharyya for short). Note that an improved algorithm [57] is presented that exploits the background information to enhance the performance of Color_Bhattacharyya; we do not compare this improved one with ours because in the current work we do not consider any background information.

For quantitative analysis, we use the the following error metric [2] called overlapping region error to compare tracking accuracy

$$
E = 1 - \frac{2|S_{resu} \cap S_{grou}|}{|S_{resu}| + |S_{grou}|}
\tag{36}
$$

where $|\cdot|$ denotes the cardinal number of a set and $S_{resu}, S_{grou}$ represent the image regions of tracking result and the ground truth, both expressed as sets of image points. The error $E$ equals 0 if the two regions are identical, increases when the overlapping region becomes small, and equals 1 if the two do not overlap at all. In our experiments, we declare the tracking failure if the error $E$ of more than five consecutive frames are larger than 0.8.

## 6.1 Qualitative and Quantitative Comparisons

The first video scenario is a very big fish tank with small cobblestones underneath. In the video clip, some fishes with pebbles on their bodies swim freely. The video clip is recorded with a hand-held camera with 430 frames (size: 352×288 pixels). The hand-held camera continuously moves; hence, the viewing angles change from time to time, causing illumination variation and background motion.

For comparison purpose, we vectorize the SIFT features as commonly done in the literature [11, 25, 27–29, 32, 34] and exploit the PCA for dimensionality reduction (henceforth PCA-SIFT). Here, we visually compare PCA-SIFT (with three dominant principal components) and Tensor-SIFT (with three component tensors). Given the reference image [the left-most image in Fig. 2(a)] of the fish, we compute its SIFT features and construct the corresponding PCA basis and tensor basis. As shown in Fig 3, for the new images available (first column), after computing their SIFT features, the corresponding features of reduced dimensionality can be achieved through the projection by PCA basis (second column) and the tensor basis matrices (third column). Although we know where the fish is, distinguishing it from the background is difficult in the PCA-SIFT images because they are so similar, whereas in the Tensor_SIFT images, the fish is distinct from the background. This is due to the PCA losing the spatial layout inherently present in the data, resulting in similar appearances of the foreground and background.

Fig. 4 illustrates in each frame the overlapping region errors of the four trackers. Before the fish swims into the background of cobblestones in frames 1–50, PCA_SIFT_EMD (yellow dotted line with "△" marker) is much better than Color_EMD (blue dashed line with "×" marker). However, afterwards, the tracking errors of PCA_SIFT_EMD (brown dotted line with "△" marker) increase sharply and become larger than those of Color_EMD. This is not surprising, because at this time, the object is nearly indistin-guishable from the cobblestone background in the PCA_SIFT images. The two trackers diverge successively at about frames 160 and 180, respectively. Color_Bhattacharyya (green dash-dotted line with "▽" marker) shrinks to some very small regions on the fish in frame 45 and never recovers. Tensor_SIFT_EMD (red solid line with "□" marker) demonstrates the best performance: it is stable and accurate in following the fish throughout the sequence. Fig. 5 shows the typical tracking results of the four trackers.

The second image sequence, a car sequence, is 300 frames long (size: 352×288 pixels); it involves a white minivan (object) on a busy road. This image sequence is also captured by a hand-held camera. At the beginning the minivan is in the shadow of the skyscrapers. It turns right, and near the zebra crossing, gradually goes out of the shadow and runs to the sunshine at about frame 60, giving rise to significant illumination variation. In frames 115 to 160, due to a dark blue minivan coming from the opposite direction, the object is occluded little by little until there is almost complete occlusion; gradually, the object becomes un-occluded once again. Afterwards, two similar situations of occlusions, which are not as severe as the first one, occur consecutively in frames 180–210 and 220–245 due to a red car and then a silver gray car passing by. Finally, the object leaves the view and disappears at about frame 285.

Fig. 6 shows the overlapping region errors of the four trackers in each frame of the car sequence. In this sequence, PCA_SIFT_EMD is better than Color_EMD in terms of both tracking accuracy and robustness. As of their respective tracking failures, the average error (mean±STD) of PCA_SIFT_EMD is 0.3845± 0.1124, whereas that of Color_EMD is 0.4224 ±0.1494. The former fails at about frame 170, whereas the latter fails at about frame 185. The tracking result of Color_ Bhatacharyya soon shrinks to some very small regions around frame 95 after the car runs into the sunshine from the skyscraper's shadow. Tensor_SIFT _EMD exhibits the best performance, following the object until its disappearance; although in the end tracking errors become large. Typical tracking results of the four trackers are shown in Fig. 9.

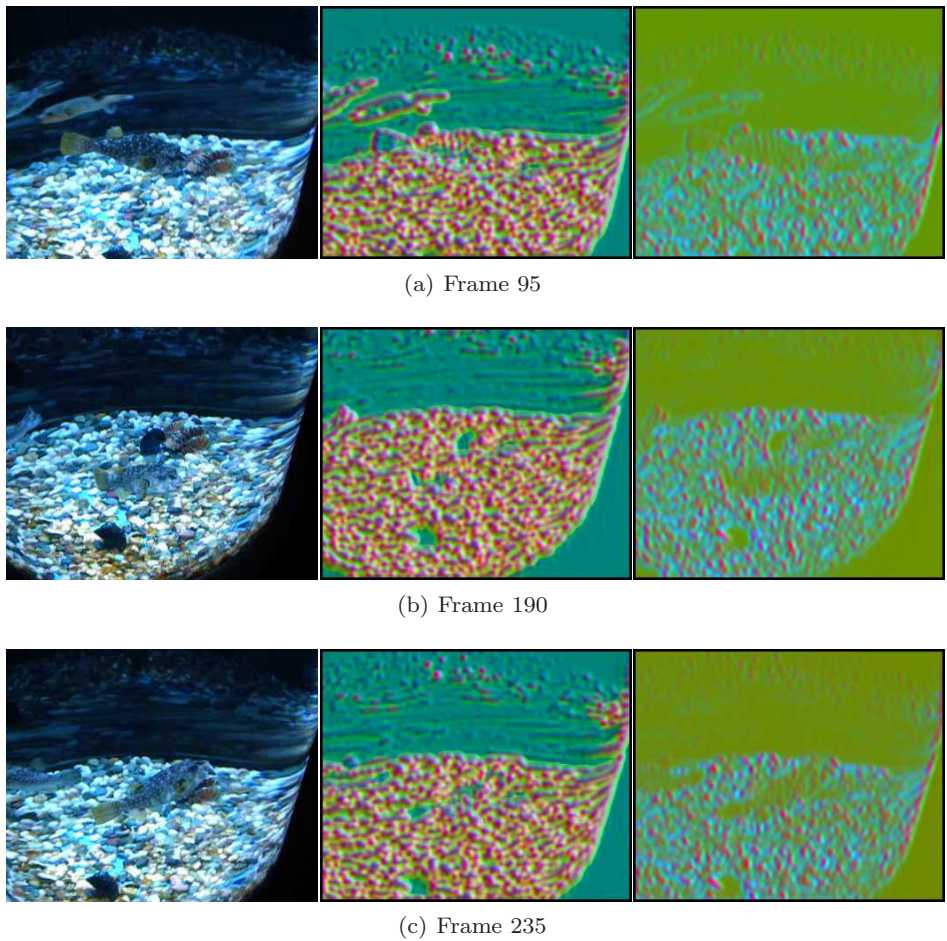

(a) Frame 95

(b) Frame 190

(c) Frame 235

**Fig. 3** Visual comparison of PCA_SIFT and Tensor_SIFT (the fish is the object). The reference image as shown in Fig. 2(a) is used to construct the PCA basis (three dominant principe components) and Tensor basis matrices (three component tensor). For the new images (left column), features of reduced dimensionality, PCA_SIFT (middle column) and Tensor_SIFT (right column), can be obtained through projection by the PCA and by the tensor decomposition, respectively. Although we know where the fish is, distinguishing it from the background is difficult in the PCA-SIFT images because they are so similar, whereas in the Tensor_SIFT image, the fish is distinct from the background.

As discussed in [28, 29], in the SIFT feature computation, color images are first transformed to gray-level images, where the SIFT features are extracted. Therefore, our Tensor-SIFT is well suited for both color and single-channel images. The third sequence (460 frames, image size: 253×288 pixels), recorded in a very crowded street, contains such single-channel (gray-level) images. In this sequence, neither Color _Bhattacharyya nor Color_EMD works, as in this case, the histograms contain inadequate information to distinguish the object from the background. In contrast, the two trackers based on SIFT features can follow the object throughout the whole image sequence. Fig. 8 shows the overlapping region errors in each frame using Tensor_SIFT_EMD (red solid line with "□" marker) and PCA_SIFT_EMD (blue dashed line with "▽" marker), respectively. Clearly, the errors of the former are generally less than those of the latter. The average errors are 0.3454 ±0.0752 for Tensor_SIFT_EMD and 0.3692±0.0772 for PCA_SIFT_EMD.

### 6.2 Discussion

The trackers that adopt the same distance measure of the EMD but depend on different features demonstrate varying performance. PCA_SIFT_EMD has bet-ter performance in the car sequence than Color_EMD in both tracking accuracy and robustness. However, in the fish sequence, the latter outperforms the former because in the PCA_SIFT images, the object (fish with pebbles on the body) and the background (cobble stones) have a similar appearance and are difficult to distinguish. In contrast, in the corresponding Tensor_SIFT images, the object is distinct from the background. Invariably, in both sequences, the tracker Tensor_SIFT_EMD shows the best performance, despite the challenging conditions of the cluttered background, illumination variation, and occlusion. The comparisons show that Tensor-SIFT is very distinctive and robust to challenging conditions.

Comparison between the two trackers, which use different distance (or similarity) measures of EMD and Bhattacharyya coefficient but the same color features, shows that Color_EMD is more robust to challenging conditions than Color_Bhatacharyya. In the fish and car sequences, Color_Bhatacharyya soon converges to very small regions on the object when illumination changes. Note that this phenomenon is also observed in the mean shift tracking that uses Bhattacharrya measure and color [7]. The contrast shows that EMD is more robust than the Bhattacharrya coefficient and is thus more competent in tracking applications in adverse environments.

Tensor-SIFT is suitable for both color and gray-level images because SIFT features are actually extracted in single-channel images [28, 29]. This is a good property because it bridges the gap between the color and gray-level images for tracking applications. Most color-based trackers fail in single-channel images, as gray-level information only does not suffice in distinguishing the foreground and background. The experiment in this sequence shows that the two trackers based on Tensor-SIFT and PCA-SIFT both works well. In addition, Tensor-SIFT is superior to PCA-SIFT in tracking accuracy.

## 7 Conclusion

This paper presents a robust method for contour tracking in adverse environments. We verified the performance of the proposed method in image sequences containing the cluttered background, illumination variation, and occlusion et al. Experiments show that the proposed method has promising performance. In summary, our primary contributions are as follows: formulation of an EMD-based functional and a two-phase method for its optimization; design of Tensor-SIFT features; and development of an integrated contour tracking algorithm.

First, we formulate a region functional based on EMD with kernel density for distribution modeling and propose a two-phase method for the functional optimization. In the first phase, computing the EMD is modeled as a transportation problem that is solved by the simplex algorithm. Next, using the theory of shape derivative, we perform a perturbation analysis around the best solution to the transportation problem, producing a PDE that governs shape evolution along the gradient descent direction of the region functional. The application of the Wasserstein distance in the one-dimensional case has been explored in active contour framework, but its application in high-dimensional cases is difficult because its closed-form does not exist [35]. To our knowledge, this work is one of the first attempts to apply the high-dimensional Wasserstein distance (EMD) in the active contour framework.

Second, we design a novel feature called Tensor-SIFT for tracking applications. We present a tensor decomposition method for dimensionality reduction of the well-known SIFT features. This SIFT-As-Tensor method can capture multiple factors inherently present in the data, i.e., two-dimensional spatial layout and phase histogram of the gradient magnitude. In contrast, the traditional method of SIFT-As-Vector, e.g., PCA, can only capture one-factor information of the histogram. Tensor-SIFT, which is applicable to both color and gray-level images, is very distinctive and insensitive to illumina-tion change and noise. These good properties plus its low dimensionality enables it to be feasible for potential applications in some other fields such as image classification and image retrieval.

Third, we develop an integrated algorithm for robust contour tracking tasks. The algorithm employs various techniques for tracking applications, e.g., the simplex algorithm, and narrow-band level set and fast marching algorithms. Particularly, we present a scheme that employs the mean shift algorithm for inter-frame contour initialization. It can provide a more accurate initial contour and thus helps PDE avoid possible local minimum during its gradient-descent process. We also introduce a realistic stopping criterion for PDE iteration. The stopping criterion is important but is no discussed broadly in the literature. Our method is simple yet effective, and is fit for automatic termination in PDE iteration.

The main problem of the proposed method is its high computational cost, as analyzed in section 5.3. To improve the computational efficiency, in future works, we will use the Graphics Processing Unit (GPU) to accomplish computational expensive procedures, including Tensor-SIFT computation, EMD calculation and PDE solution. We are also interested in further enhancing the tracker's performance, by combining multiple image information such as background characteristics or optical flow.

## References

1. Aubert, G., M. Barlaud, O. Faugeras, and S. Jehan-Besson (2003). Image segmentation using active contours: Calculus of variations or shape gradients? *SIAM Applied Mathematics 63*(6), 2128–2154.

2. Bajramovic, F., C. Grabl, and J. Denzler (2005). Efficient combination of histograms for real-time tracking using mean-shift and trust-region optimization. In *Proc. 27th DAGM Symposium on Pattern Recognition*, pp. 254–261.

3. Blake, A. and M. Isard (1998). *Active contours: The Application of Techniques from Graphics, Vision, Control Theory and Statistics to Visual Tracking of Shapes in Motion*. Springer-verlag.

4. Brox, T., M. Rousson, R. Deriche, and J. Weickert (2010, March). Colour, texture, and motion in level set based segmentation and tracking. *Image Vision Comput. 28*, 376–390.

5. Caselles, V., R. Kimmel, and G. Sapiro (1997). Geodesic active contours. *Int. J. of Comput. Vision 22*(1), 61–79.

6. Chan, T. F. and L. A. Vese (2001). Active contours without edges. *IEEE Trans. on Image Processing 10*(2), 266–277.

7. Collins, R. T. (2003). Mean-shift blob tracking through scale space. In *Proc. IEEE Conf. Comp. Vis. Patt. Recog.*, pp. 234–241.

8. Comaniciu, D., V. Ramesh, and P. Meer (2000). Real-time tracking of non-rigid objects using mean shift. In *Proc. IEEE Conf. Comp. Vis. Patt. Recog.*, pp. 142–149.

9. Comaniciu, D., V. Ramesh, and P. Meer (2003). Kernel-based object tracking. *IEEE Trans. Pattern Anal. Mach. Intell. 25*(5), 564–575.

10. Cremers, D., M. Rousson, and R. Deriche (2007, April). A review of statistical approaches to level set segmentation: integrating color, texture, motion and shape. *Int. J. of Comput. Vision 72*(2), 195–215.

11. Csurka, G., C. Bray, C. Dance, and L. Fan (2004). Visual categorization with bags of keypoints. In *ECCV Workshop on Statistical Learning in Computer Vision*, pp. 1–22.

12. Delfour, M. C. and J.-P. Zolesio (2001). *Shapes and Geometries: Analysis, Differential Calculus, and Optimization*. SIAM.

13. Faber, N. K. M., R. Bro, and P. K. Hopke (2003). Recent developments in cande-comp/parafac algorithms: A critical review. *Chemometrics and Intelligent Laboratory Systems 65*, 119–137.

14. Fitzgibbon, A., M. Pilu, and R. B. Fisher (1999). Direct least square fitting of ellipses. *IEEE Trans. Pattern Anal. Mach. Intell. 21*(5), 476–480.

15. Fleet, D. J. and Y. Weiss (2006). *Handbook of Mathematical Models in Computer Vision, ch. 15*, Chapter Optical Flow Estimation. Springer-Verlag.

16. Freedman, D. and T. Zhang (2004). Active contours for tracking distributions. *IEEE Trans. on Image Processing 13*(4), 518–526.

17. Haker, S., L. Zhu, A. Tannenbaum, and S. Angenent (2004, December). Optimal mass transport for registration and warping. *Int. J. of Comput. Vision 60*, 225–240.

18. Herbulot, A., S. Jehan-Besson, S. Duffner, M. Barlaud, and G. Aubert (2006). Segmentation of vectorial image features using shape gradients and information measures. *J. Math. Imaging Vis. 25*(3), 365–386.

19. Irani, M. and P. Anandan (1998). A unified approach to moving object detection in 2d and 3d scenes. *IEEE Trans. Pattern Anal. Mach. Intell. 20*(6), 577–589.

20. Jehan-Besson, S., M. Barlaud, G. Aubert, and O. Faugeras (2003). Shape gradients for histogram segmentation using active contours. In *Proc. of the Int. Conf. on Computer Vision*, Washington, DC, USA, pp. 408. IEEE Computer Society.

21. Kantorovich, L. V. (1942). On the translocation of masses. *Akademii Nauk SSSR 37*, 199–201.

22. Kass, M., A. P. Witkin, and D. Terzopoulos (1988). Snakes: Active contour models. *Int. J. of Comput. Vision 1*(4), 321–331.

23. Kolda, T. G. and B. W. Bader (2009, September). Tensor decompositions and applications. *SIAM Review 51*(3), 455–500.

24. Lazebnik, S., C. Schmid, and J. Ponce (2006). Beyond bags of features: Spatial pyramid matching for recognizing natural scene categories. In *Proc. IEEE Conf. Comp. Vis. Patt. Recog.*, Washington, DC, USA, pp. 2169–2178. IEEE Computer Society.

25. Li, F.-F. and P. Perona (2005). A bayesian hierarchical model for learning natural scene categories. In *Proc. IEEE Conf. Comp. Vis. Patt. Recog.*, Washington, DC, USA, pp. 524–531. IEEE Computer Society.

26. Ling, H. and K. Okada (2007). An efficient earth mover's distance algorithm for robust histogram comparison. *IEEE Trans. Pattern Anal. Mach. Intell. 29*(5), 840–853.

27. Liu, C., J. Yuen, A. Torralba, J. Sivic, and W. T. Freeman (2008). Sift flow: Dense correspondence across different scenes. In *Proc. of European Conf. on Computer Vision*, Berlin, Heidelberg, pp. 28–42. Springer-Verlag.

28. Lowe, D. G. (1999). Object recognition from local scale-invariant features. In *Proc. of the Int. Conf. on Computer Vision*, Washington, DC, USA, pp. 1150. IEEE Computer Society.

29. Lowe, D. G. (2004). Distinctive image features from scale-invariant keypoints. *Int. J. of Comput. Vision 60*(2), 91–110.

30. Luenberger, D. G. and Y. Ye (2007). *Linear and Nonlinear Programming*. Springer-Verlag.

31. McKenna, S. J., Y. Raja, and S. Gong (1999). Tracking colour objects using adaptive mixture models. *Image Vision Comput. 17*(3-4), 225–231.

32. Mikolajczyk, K. and C. Schmid (2005). A performance evaluation of local descriptors. *IEEE Trans. Pattern Anal. Mach. Intell. 27*(10), 1615–1630.

33. Monge, G. (1781). Mémoire sur la théorie des déblais et des remblais. *Hist. de l'Acad. des Sciences de Paris*, 666–704.

34. Mutch, J. and D. G. Lowe (2006, June). Multi-class object recognition with sparse, localized features. In *Proc. IEEE Conf. Comp. Vis. Patt. Recog.*, New York, NY, pp. 11–18.

35. Ni, K., X. Bresson, T. Chan, and S. Esedoglu (2009). Local histogram based segmentation using the wasserstein distance. *Int. J. of Comput. Vision 84*(1), 97–111.

36. Osher, S. and R. Fedkiw (2002). *Level Set Methods and Dynamic Implicit Surfaces*. Springer-Verlag.

37. Osher, S. and J. A. Sethian (1988). Fronts propagating with curvature-dependent speed: algorithms based on hamilton-jacobi formulations. *J. Comput. Phys. 79*(1), 12–49.

38. Paragios, N. and R. Deriche (2000). Geodesic active contours and level sets for the detection and tracking of moving objects. *IEEE Trans. Pattern Anal. Mach. Intell. 22*(3), 266–280.

39. Paragios, N. and R. Deriche (2005). Geodesic active regions and level set methods for motion estimation and tracking. *Comput. Vis. Image Underst. 97*(3), 259–282.

40. Parlett, B. N. (1998). *The symmetric eigenvalue problem.* Upper Saddle River, NJ, USA: Prentice-Hall, Inc.

41. Pele, O. and M. Werman (2008). A linear time histogram metric for improved sift matching. In *Proc. of Eur. Conf. Comp. Vis.*, pp. 495–508.

42. Pele, O. and M. Werman (2009). Fast and robust earth mover's distances. In *Proc. Int. Conf. Comp. Vis.*

43. Peleg, S., M. Werman, and H. Rom (1989). A unified approach to the change of resolution: Space and gray-level. *IEEE Trans. Pattern Anal. Mach. Intell. 11*(7), 739–742.

44. Precioso, F., M. Barlaud, T. Blu, and M. Unser (2005, July). Robust real-time segmentation of images and videos using a smooth-spline snake-based algorithm. *IEEE Trans. on Image Processing 14*(7), 910–924.

45. Rachev, S. and L. Rüschendorf (1998). *Mass transportation problems. Vol. I: Theory, Vol. II: Applications. Probability and Its Application.* Springer-verlag.

46. Rathi, Y., N. Vaswani, A. Tannenbaum, and A. Yezzi (2007). Tracking deforming objects using particle filtering for geometric active contours. *IEEE Trans. Pattern Anal. Mach. Intell. 29*(8), 1470–1475.

47. Ronfard, R. (1994). Region-based strategies for active contour models. *Int. J. of Comput. Vision 13*(2), 229–251.

48. Rubner, Y., C. Tomasi, and L. J. Guibas (2000). The earth mover's distance as a metric for image retrieval. *Int. J. of Computer Vision 40*(2), 99–121.

49. Sethian, J. (1999). *Level Set Methods and Fast Marching Methods.* Cambridge University Press.

50. Sokolowski, J. and J.-P. Zolésio (1992). *Introduction to shape optimization. Shape sensitivity analysis.* Springer Ser. Comput. Math. 16. Springer-Verlag.

51. Turk, M. and A. Pentland (1991). A unified approach to the change of resolution: Space and gray-level. *Journal of Cognitive Neuroscience 3*(1), 71–862.

52. Vasilescu, M. and D. Terzopoulos (2002). Multilinear analysis of image ensembles: Tensorfaces. In *Proc. of European Conf. on Computer Vision*, Berlin, Heidelberg, pp. 447C460. Springer-Verlag.

53. Vasilescu, M. and D. Terzopoulos (2003). Multilinear subspace analysis for image ensembles. In *Proc. of IEEE Conf. on Computer Vision and Pattern Recognition*, Washington, DC, USA, pp. 93–99. IEEE Computer Society.

54. Wang, H. and N. Ahuja (2008). A tensor approximation approach to dimensionality reduction. *Int. J. of Comput. Vision 76*(3), 217–229.

55. Wu, C. (2007). SiftGPU: A GPU implementation of scale invariant feature transform (SIFT). http://cs.unc.edu/~ccwu/siftgpu.

56. Yan, K. and R. Sukthankar (2004). PCA-SIFT: a more distinctive representation for local image descriptors. In *Proc. IEEE Conf. Comp. Vis. Patt. Recog.*, pp. 506–513. IEEE Computer Society.

57. Zhang, T. and D. Freedman. (2005). Improving performance of distribution tracking through background mismatch. *IEEE Trans. Pattern Anal. Mach. Intell. 27*(2), 282–287.

58. Zhao, Q., Z. Yang, and H. Tao (2010). Differential earth mover's distance with its applications to visual tracking. *IEEE Trans. Pattern Anal. Mach. Intell. 32*(2), 274–287.

59. Zhong, Y., A. K. Jain, and M.-P. Dubuisson-Jolly (2000). Object tracking using deformable templates. *IEEE Trans. Pattern Anal. Mach. Intell. 22*(5), 544–549.

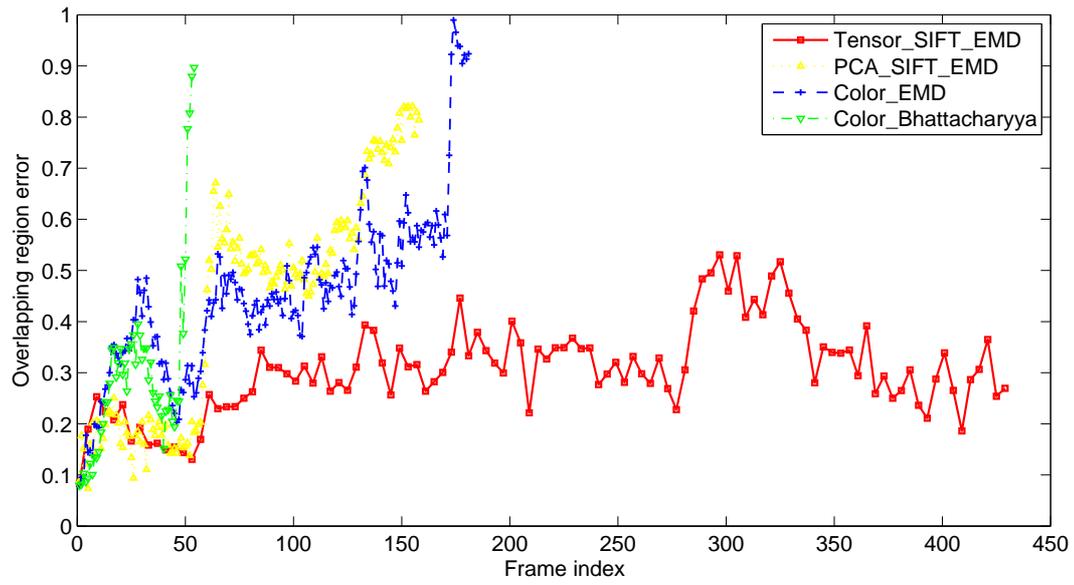

**Fig. 4** Overlapping region errors of the four trackers in the fish sequence.

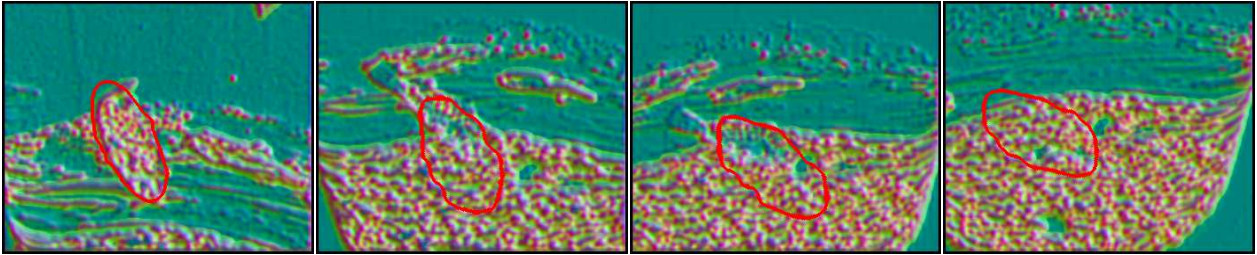

(a) Results using PCA_SIFT_EMD in the PCA_SIFT images. Tracking fails at about frame 160; from left to right are frames 15, 60, 90 and 150.

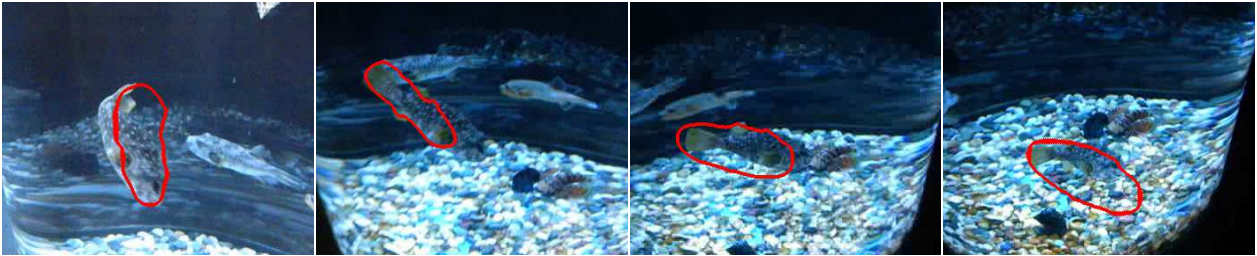

(b) Results using Color_EMD in the original color images. Tracking fails at about frame 180; from left to right are frames 15, 60, 95 and 160.

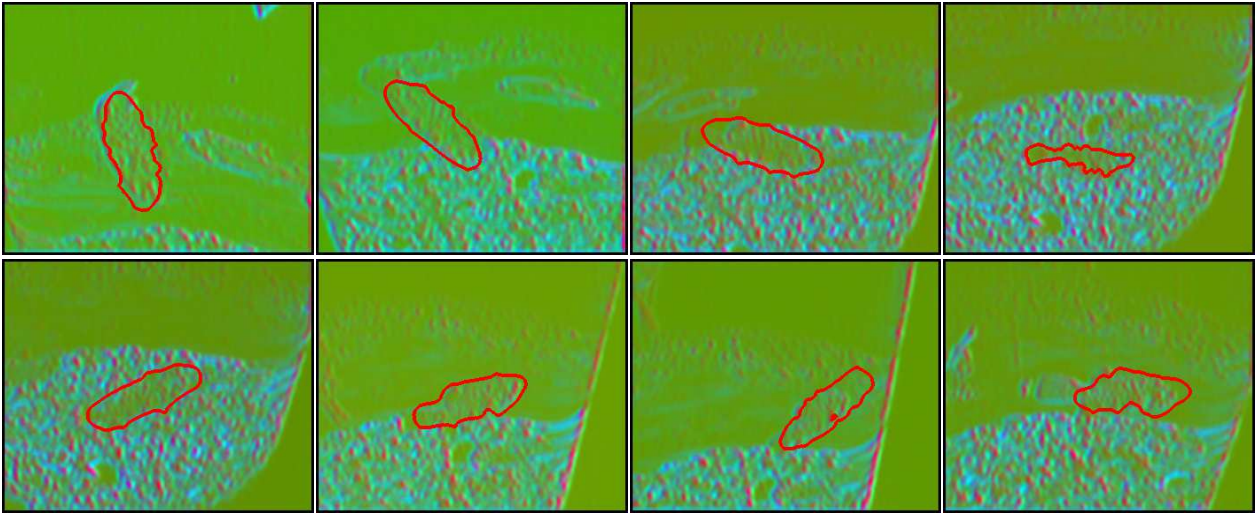

(c) Results using Tensor_SIFT_EMD in the Tensor_SIFT images. The tracker successfully follows the object throughout the sequence; from left-top to right-bottom are frames 15, 60, 95, 160, 220, 270, 300 and 375.

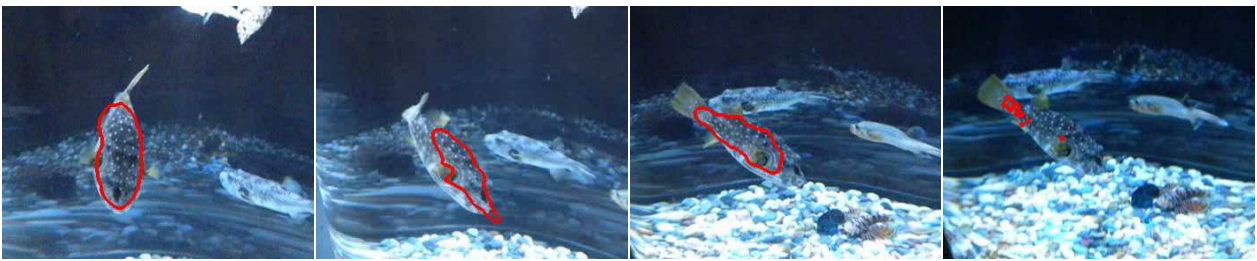

(d) Results using Color_Bhatacharyya in the original color images. Tracking fails at about frame 45; from left to right are frames 1, 20, 40 and 55.

**Fig. 5** Typical tracking results in the fish sequence using the four trackers.

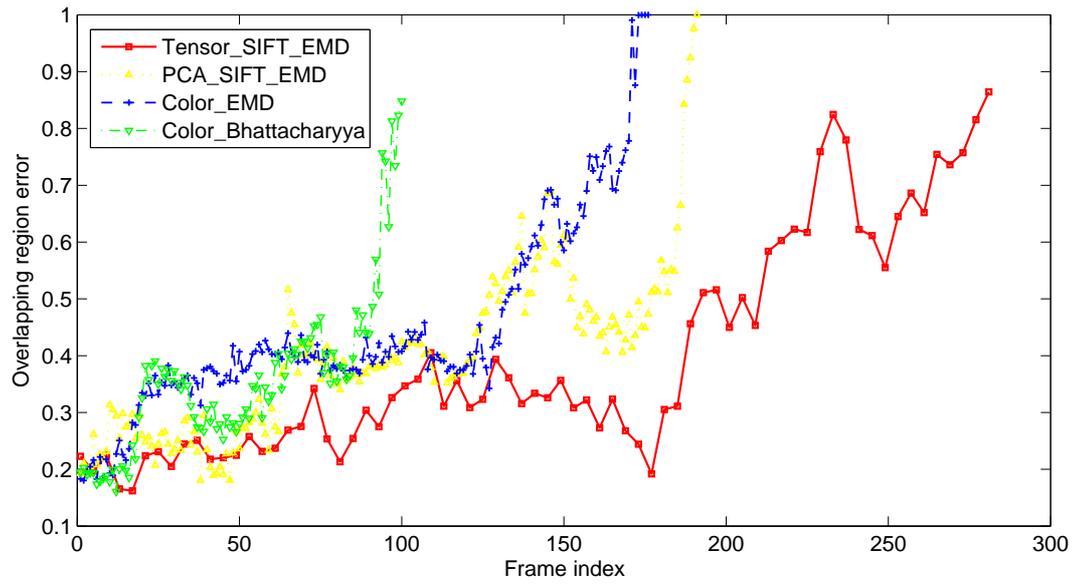

**Fig. 6** Overlapping region errors of four trackers in the car sequence.

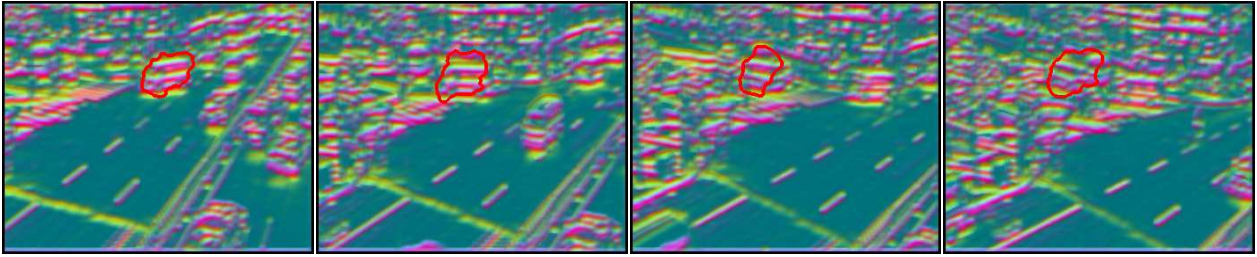

(a) Results using PCA_SIFT_EMD in the PCA_SIFT images. Tracking fails at about frame 170; from left to right are frames 20, 70, 120 and 165.

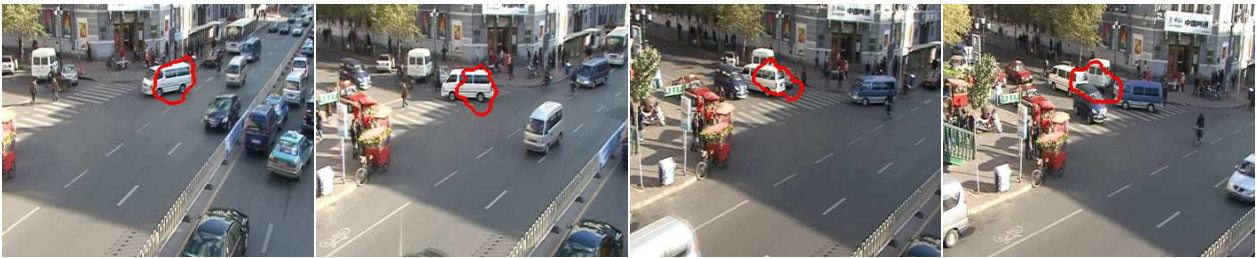

(b) Results using Color_EMD in the original color images. Tracking fails at about frame 185; from left to right are frames 20, 70, 120 and 165.

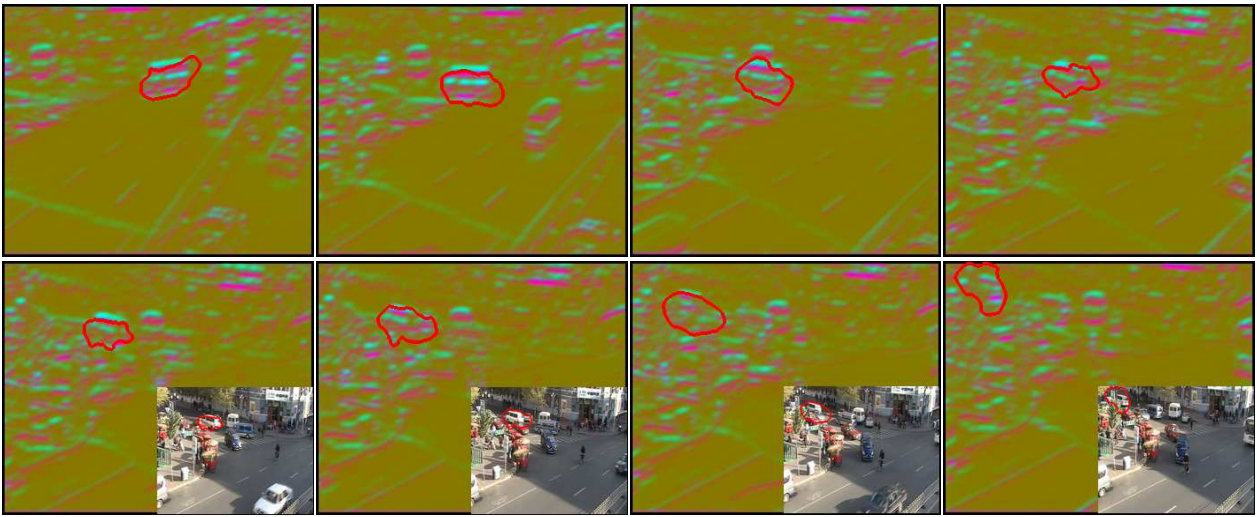

(c) Results using Tensor_SIFT_EMD in the Tensor_SIFT images. The tracker manages to follow the object until its disappearance; from left-top to right-bottom are frames 20, 70, 120, 165, 185, 205, 240 and 260. The small images imposed in the second row show the corresponding color scenes.

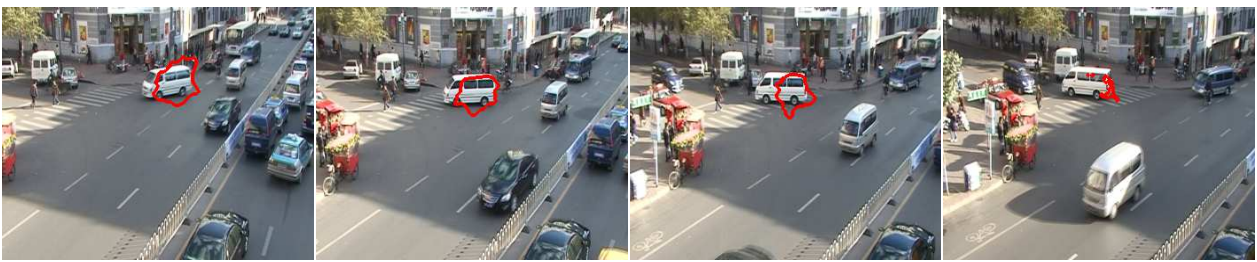

(d) Results using Color_Bhattacharyya in the original color images. Tracking fails at about frame 95; from left to right are frames 20, 45, 70 and 95.

**Fig. 7** Typical tracking results in the car sequence using the four trackers.

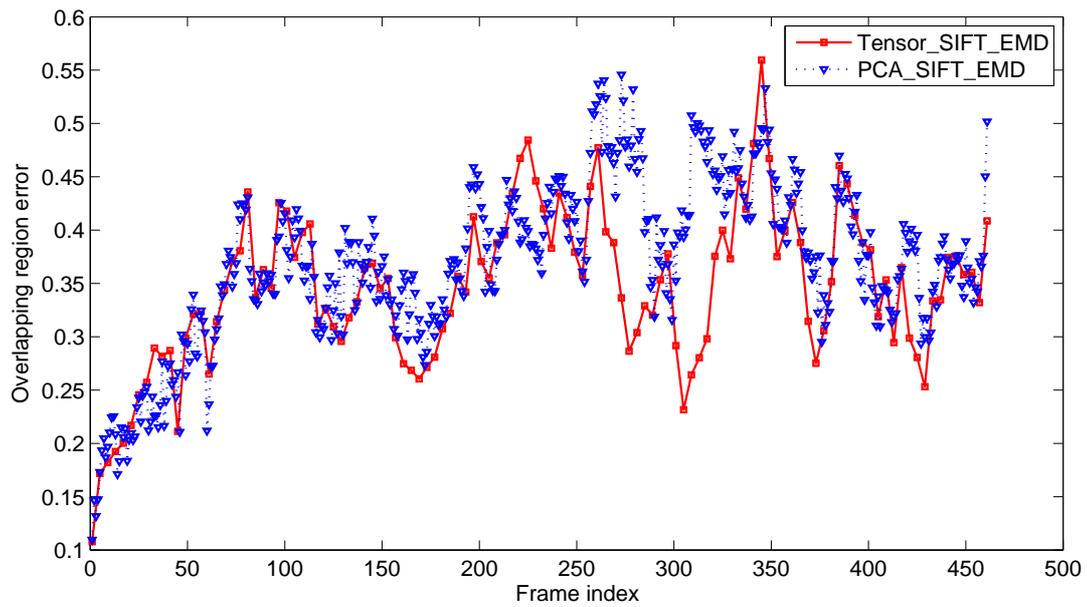

**Fig. 8** Overlapping region errors of Tensor_SIFT_EMD and PCA_SIFT_EMD in the gray-level pedestrian sequence. Their average errors (mean±STD) are 0.3454±0.0752 and 0.3692±0.0772, respectively.

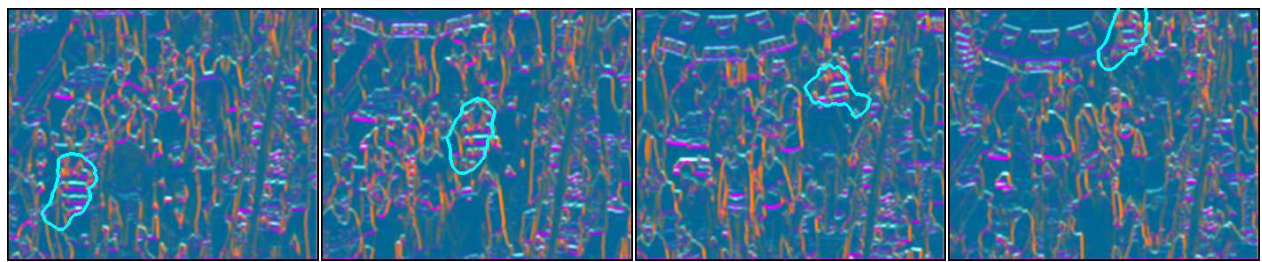

(a) Results using PCA_SIFT_EMD in the PCA_SIFT images. From left to right are frames 40, 150, 290 and 450.

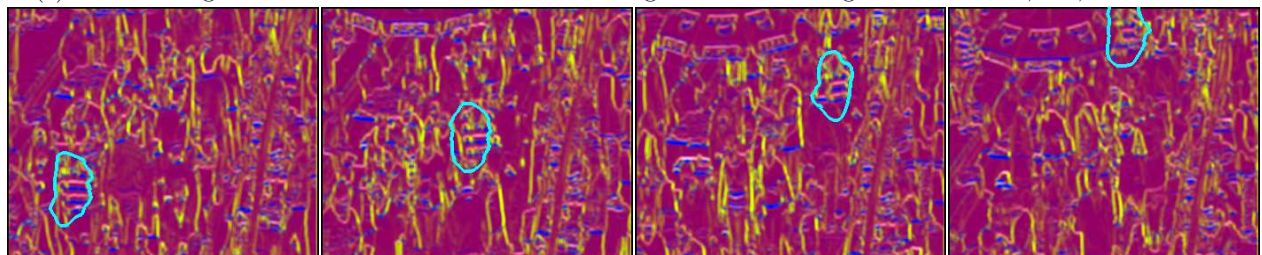

(a) Results using Tensor_SIFT_EMD in the Tensor_SIFT images. From left to right are frames 40, 150, 290 and 450.

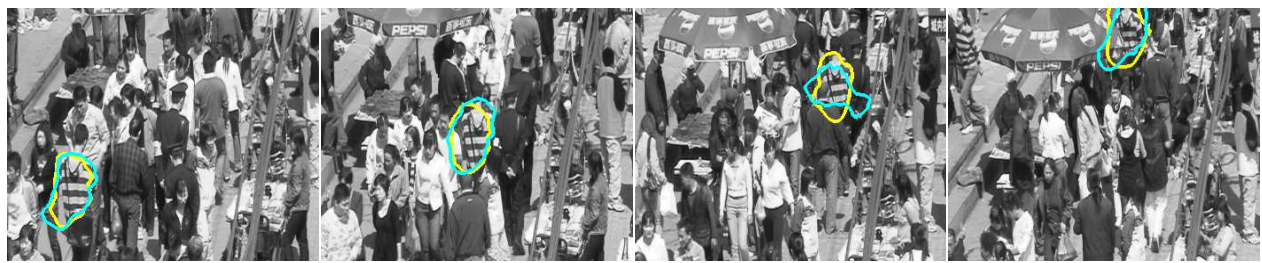

(b) Results of Tensor_SIFT_EMD (brown solid lines) and PCA_SIFT_EMD (cyan solid lines) drawn in the original gray-level images. From left to right are frames 40, 150, 290 and 450.

**Fig. 9** Typical tracking results in the single-channel gray-level pedestrian sequence.